\documentclass[journal]{IEEEtran}
\ifCLASSINFOpdf
\else
\fi

\usepackage{xcolor}
\usepackage{booktabs}
\usepackage{epsfig}
\usepackage{graphicx}
\usepackage{amsmath}

\usepackage{amssymb}

\usepackage{verbatim}

\usepackage{url}
\usepackage{tabularx}
\usepackage{multirow}
\usepackage{subfigure}
\usepackage{amsmath,dsfont}
\usepackage{array}
\usepackage{enumitem}

\usepackage{pifont}
\usepackage{threeparttable}
\usepackage{bbding}

\usepackage{verbatim}

\usepackage{colortbl}
\definecolor{mygray}{gray}{.75}

\usepackage{xcolor}

\usepackage[normalem]{ulem}

\newcommand{\etal}{\emph{et al.}~}
\newcommand{\ie}{\emph{i.e.,}~}

\usepackage{caption}

\begin{document}
%

\title{Reading-strategy Inspired Visual Representation Learning for Text-to-Video Retrieval}
%
%
%

\author{Jianfeng~Dong,
        Yabing~Wang,
        Xianke~Chen,
        Xiaoye~Qu,
        Xirong~Li,~\IEEEmembership{Member,~IEEE,}
        Yuan~He,
        and~Xun~Wang,~\IEEEmembership{Member,~IEEE}
\thanks{Manuscript received November 5, 2021; revised January 14, 2022; Accepted February 8. This work was supported by National Key R\&D Program of China (No.~2018YFB1404102), NSFC (No.~62172420, No.~61902347, No. 61976188), the Public Welfare Technology Research Project of Zhejiang Province (No.~LGF21F020010), the Research Program of Zhejiang Lab (No.~2019KD0AC02), the Open Projects Program of National Laboratory of Pattern Recognition, and the Fundamental Research Funds for the Provincial Universities of Zhejiang.
The article was recommended by Associate Editor Dr. H. Huang.
(Jianfeng Dong and Yabing Wang are co-first authors. Corresponding authors: Xirong Li and Xun Wang)}
\thanks{J. Dong is with the College of Computer and Information Engineering, Zhejiang Gongshang University, Hangzhou 310035, China, and also with the State Key Laboratory of Information Security (Institute of Information Engineering, Chinese Academy of Sciences, Beijing 100093). E-mail: dongjf24@gmail.com}
\thanks{Y. Wang, X. Chen and X. Wang are with the College of Computer and Information Engineering, Zhejiang Gongshang University, Hangzhou 310035, China.
E-mail: wyb7wyb7@163.com}
\thanks{X. Qu is with the School of Electronic Information and Communications, Huazhong University of Science and Technology, Hubei 430074, China. E-mail: xiaoye@hust.edu.cn}
\thanks{X. Li is with the Key Lab of Data Engineering and Knowledge Engineering, Renmin University of China, and the AIMC Lab, School of Information, Renmin University of China, Beijing 100872, China. E-mail: xirong@ruc.edu.cn}
\thanks{Y. He is with the Alibaba Group, Beijing 100102, China. E-mail: heyuan.hy@alibaba-inc.com}
}

%
%

\markboth{IEEE Transactions on Circuits and Systems for Video Technology,~Vol.~x, No.~x,~2022}
{Shell \MakeLowercase{\textit{et al.}}: Bare Demo of IEEEtran.cls for IEEE Journals}
%



\maketitle

\begin{abstract}
 This paper aims for the task of text-to-video retrieval, where given a query in the form of a natural-language sentence, it is asked to retrieve videos which are semantically relevant to the given query, from a great number of unlabeled videos. The success of this task depends on cross-modal representation learning that projects both videos and sentences into common spaces for semantic similarity computation. In this work, we concentrate on video representation learning, an essential component for text-to-video retrieval. Inspired by the reading strategy of humans, we propose a Reading-strategy Inspired Visual Representation Learning (RIVRL) to represent videos, which consists of two branches: a previewing branch and an intensive-reading branch. The previewing branch is designed to briefly capture the overview information of videos, while the intensive-reading branch is designed to obtain more in-depth information. Moreover, the intensive-reading branch is aware of the video overview captured by the previewing branch. Such holistic information is found to be useful for the intensive-reading branch to extract more fine-grained features. Extensive experiments on three datasets are conducted, where our model RIVRL achieves a new state-of-the-art on TGIF and VATEX. Moreover, on MSR-VTT, our model using two video features shows comparable performance to the state-of-the-art using seven video features and even outperforms models pre-trained on the large-scale HowTo100M dataset. Code is available at \url{https://github.com/LiJiaBei-7/rivrl}.
\end{abstract}

\begin{IEEEkeywords}
Cross-modal retrieval, video-text retrieval, video representation learning, preview-aware attention.
\end{IEEEkeywords}

%
\IEEEpeerreviewmaketitle

\section{Introduction}\label{sec:introduction}

\IEEEPARstart{W}{ith} the increasing popularity of video streaming platforms such as YouTube and TikTok, there has been an explosive growth of video data on the Internet. Naturally, there is an increasing need of automated methods for recognizing~\cite{huang2021holographic,wu2021spatiotemporal,tan2021selective,li2021interventional}, describing \cite{xu2018dual,deng2021syntax,shang2019annotating,dong2016early} and retrieving \cite{araujo2017large,gao2021learning,yang2021deconfounded,yang2022video} the video content. 
In this paper, we focus on text-to-video retrieval. Given a query in the form of a natural-language sentence, the task is to retrieve videos semantically relevant to the given query from many unlabeled videos.

For building such a video retrieval model, how to compute the semantic similarity between two modalities, \ie text and videos, is crucial.
Earlier efforts for text-to-video retrieval are concept-based methods~\cite{cikm13-zsvr,aaai15-zsed,icmr2017-certh-avs,pami2017-videostory}, which represent videos and textual queries into a pre-defined concept space, and the similarity is computed by concept matching.
As the limited performance of concept-based methods,  cross-modal representation learning based methods are preferred, which learns a joint embedding space in a concept-free manner for cross-modal similarity measurement and shows better performance~\cite{wang2020learning,song2021spatial,yang2020weakly,wang2021t2vlad,he2021improving,xiao2020visual,dong2019dual}.

\begin{figure}[t!]
\centering
\includegraphics[width=0.99\columnwidth]{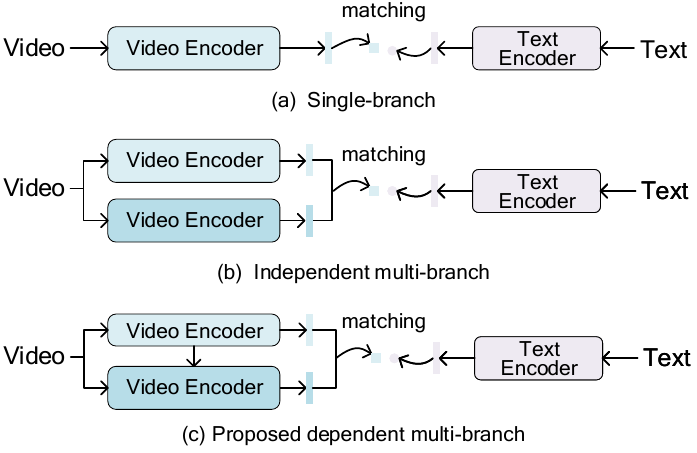}
\caption{Three visual representation learning paradigms for text-to-video matching and retrieval: (a) Single-branch, (b) Independent multi-branch, and (c) Proposed dependent multi-branch visual representation learning.
} \label{fig:title_fig}\vspace{-4mm}
\end{figure}

Following the line of cross-modal representation learning based method, in this work we concentrate on video representation learning, an essential component for text-to-video retrieval.
A typical approach of video representation learning is to first extract visual features from video frames by pre-trained CNN models, and subsequently aggregate the frame-level features into a video-level feature by mean pooling~\cite{eccv2016ws-otani,mithun2018learning} or max pooling~\cite{wray2019fine,miech2019howto100m}. A Fully Connected (FC) layer is usually further employed to map the video-level feature into a joint embedding space. Instead of the simple pooling strategies, sequence-aware deep neural networks were explored. For instance,  Long Short-Term Memory (LSTM) in~\cite{torabi2016learning}, Gated Recurrent Unit (GRU) in~\cite{dongdl}. Besides, GRU stacked with a multi-head self-attention layer was also investigated to represent videos~\cite{sigir2020tree}.
Transformer~\cite{vaswani2017attention} based methods were also proposed for video representation learning~\cite{gabeur2020multi,zhu2020actbert}.
However, the above methods extract video features by a single-branch structure, as illustrated in Fig. \ref{fig:title_fig} (a), which may not capture the video information comprehensively.

Recently, we noticed an increasing use of employing a multi-branch structure to represent videos~\cite{miech2018learning,song2021spatial,li2020sea,chen2020fine,song2019polysemous,wang2021t2vlad}, as shown in Fig \ref{fig:title_fig} (b). For instance, Miech \etal \cite{miech2018learning} used multiple distinct video features and employed a gated embedding module to learn a new feature representation for each input feature.
With the same input of each branch, Li \etal \cite{li2020sea} utilized multiple parallel FC layers to project the input into multiple distinct video representation spaces, wherein multiple branches were independent. More recently, Wang \etal \cite{wang2021t2vlad} proposed to process a given video by a global branch and a local branch, where two branches were also independent without any interactions.
Despite their state-of-the-art performance, we argue that such an independent multiple-branch paradigm is suboptimal. 
According to our observations, when the two branches learn independently, both branches tend to learn main objects in videos while ignoring specific key details.
Therefore, in order to strengthen the complementarity between multiple branches, it is necessary to design a representation learning network with a distinct granularity for each branch, while exploring the cross-branch association learning mechanism.

To alleviate the limitation mentioned above, we propose a Reading-strategy Inspired Visual Representation Learning (RIVRL) model.
We are inspired by the reading strategy of humans, where humans typically preview text with a quick glance to obtain an overview of the text, and then intensively read it with the text overview to gain a deeper understanding.
Our hypothesis is that if one model first briefly knows the video overview, it will help the model understand the video more deeply.
Therefore, our proposed video representation model consists of two branches: a previewing branch and an intensive-reading branch, as illustrated in Fig \ref{fig:title_fig} (c).
The previewing branch is implemented as a lightweight video encoder, which is designed to briefly capture the overview information of videos. By contrast, the intensive-reading branch is implemented as a relatively heavy video encoder, which is designed to obtain more in-depth information.
The key point is that the earlier generated representation from the previewing branch is then integrated into the representation learning of the intensive-reading branch. 
So two branches are dependent.
Besides, the representation of previewing branch to some extent can be regarded as the overview of the given video.
In this way, the intensive-reading branch is aware of the video overview, which typically helps it extract more fine-grained features. 

Not surprisingly, the proposed method uses existing technologies including~\cite{cho2014learning}, hierarchical modeling~\cite{ging2020coot}, hybrid space learning~\cite{dong2021dual}, \textit{etc}. Meanwhile, the proposed model architecture can be viewed as hierarchical at a high level. Our novelty is two-fold. First, we propose a dependent two-branch network for video representation learning. To the best of our knowledge, existing methods for text-to-video retrieval either use a single branch~\cite{sigir2020tree,dong2021dual,luo2021coco}, or use multiple but independent branches~\cite{wang2021t2vlad,chen2020fine,liu2019use}. Second, to effectively model such dependency, we develop a new feature aggregation module termed Previewing-aware Attention (PaA). Different from the self-attention module used in Transformers~\cite{vaswani2017attention}, where the same input feature sequence is used as queries (Q), keys (K), and values (V), the Q of PaA is taken from the previewing branch, and thus not self-attended. These two novel elements are shown to be important for learning a comprehensive visual representation for text-to-video retrieval.

To sum up, the contributions of this paper are as follows:
\begin{itemize}
    \item We propose Reading-strategy Inspired Visual Representation Learning (RIVRL), where a previewing branch and an intensive-reading branch are jointly learned to extract cross-modal video representations. The intensive-reading branch, being aware of the representation learned by the previewing branch, can adaptively extract more fine-grained information.
    \item In the intensive-reading branch, we devise a multi-granularity segment representation and a previewing-aware attention. All these components are beneficial to the text-to-video retrieval task.
    \item We conduct extensive experiments on three challenging video datasets, \ie MSR-VTT~\cite{xu2016msr}, TGIF~\cite{tgif} and VATEX~\cite{wang2019vatex}. The proposed model achieves a new state-of-the-art (SOTA) on TGIF and VATEX. While on MSR-VTT where the SOTA \cite{wang2021t2vlad} uses seven video features, our model obtains comparable performance with just two 2D-CNN features.  
\end{itemize}

\section{Related Work} \label{sec:rel-work}

As this work focuses on video representation for text-to-video retrieval, we mainly review recent progress in this direction.  Depending on whether multi-branch networks are used for video representation,  we categorize existing methods into two groups, \ie single-branch methods \cite{torabi2016learning,miech2019howto100m,zhu2020actbert,feng2020exploiting} and multiple-branch methods \cite{song2021spatial,li2020sea,liu2019use,arandjelovic2016netvlad}.
Besides, we also review the related works in terms of the cross-modal similarity computation.

\subsection{Single-branch Methods}
A common implementation of single-branch methods was to first extract visual features from video frames by pre-trained CNN models, and subsequently aggregate the frame-level features into a video-level feature. To aggregate the features, mean pooling \cite{dong2018predicting,shao2018find} and max pooling \cite{wray2019fine,miech2019howto100m} were the popular choices, as they were very simple and parameter-free.
However, they ignored the temporal order in videos which was typically essential for video representation.
Hence, Torabi \etal \cite{torabi2016learning} proposed to use an LSTM to explicitly model the temporal dependency information, where frame-level features were sequentially fed into the LSTM and the mean pooling of the hidden vectors at each step was used as the video representation.
Instead of LSTM, Dong \etal \cite{dongdl} employed GRU considering it was more lightweight, and utilized a simple self attention that was implemented by an FC layer to aggregate features. With the similar idea of~\cite{dongdl}, Yang \etal \cite{sigir2020tree} also utilized the GRU for temporal modeling, but additionally used a multi-head self-attention layer~\cite{vaswani2017attention} to learn the frame-wise correlation thus obtaining the stronger video representation.
Different from the above methods that learned the video representation based frame-level features extracted by CNN models, Feng \etal
\cite{feng2020exploiting} proposed to employ an object detector model to extract object-level features, and used a Graph Convolutional Network (GCN) to model the semantic relations between objects thus obtained more fine-grained video representation. 
Recently, with the success of Transformer in Natural Language Processing (NLP), we observed an increasing use of such techniques for video representation in the task of text-to-video retrieval~\cite{gabeur2020multi,zhu2020actbert,li2020hero,luo2021coco}.
For instance, Gabeur \etal~\cite{gabeur2020multi} proposed a model called multi-modal Transformer (MMT) with four stacked Transformer layers, which jointly encoded seven diverse video features for video representation. In \cite{zhu2020actbert}, Zhu and Yang proposed ActBERT to jointly encode global actions, local regional objects, and text descriptions in a BERT-like Transformer framework. 
In~\cite{luo2021coco}, Luo \etal also proposed a BERT-like encoder-decoder structure called CoCo-BERT, which simultaneously pursued inter-modal matching and intra-modal denoising between masked and unmasked inputs in a contrastive manner.

\subsection{Multiple-branch Methods}

For multiple-branch methods, multiple parallel video encoding branches were jointly used to represent videos.
One simple way was to utilize multiple independent encoding branches with different video features as inputs~\cite{miech2018learning,liu2019use,song2021spatial,he2021improving}.
In~\cite{miech2018learning}, for each branch, Miech \etal first aggregated input features into video-level feature vectors by mean pooling, max pooling or NetVLAD, then a gated embedding module was further employed for feature mapping. They adopted seven different features, and seven different mapped features obtained from seven branches were used for video representation. In the follow-up work, Liu \etal \cite{liu2019use} proposed a model named Collaborative Experts (CE), where they also employed seven branches for video representation but used a collaborative gating to fuse the outputs of multiple branches.
Recently, Song \etal \cite{song2021spatial} devised two separated branches with distinct structures for video representation, where one branch used a GCN to model spatial-temporal relations between objects detected in video frames, the other employed mean pooling over the frame-level frames.
Another popular way was devising multiple encoding branches with the same video feature as input~\cite{li2020sea,ging2020coot,chen2020fine,wang2021t2vlad,song2019polysemous,chen2021learning}.
For instance, Li \etal \cite{li2020sea} proposed Sentence Encoder Assembly (SEA), where the authors utilized multiple different text encoders. Besides, they utilized multiple parallel FC layers over the mean-pooled frame-level feature for feature mapping, where each video branch was corresponding to a specific text encoder.
In \cite{dong2021dual}, Dong \etal proposed Dual Encoding for text-to-video retrieval, where they devised a multi-level video encoding including mean pooling, biGRU and CNN for video representation. 
In \cite{chen2020fine}, Chen \etal proposed a model called Hierarchical Graph Reasoning~(HGR). They devised three branches to capture three hierarchical semantic levels in videos, which were responsible to capture global events, local actions and entities, respectively.
As the follow-up work of \cite{chen2020fine}, Chen \etal proposed Generalized Pooling Operator~(GPO)~\cite{chen2021learning}, where the authors plugged the generalized pooling operator as the video and text feature aggregator in HGR.
The works~\cite{song2019polysemous,ging2020coot,wang2021t2vlad} shared the similar idea of using two branches to learn global and local information in videos respectively.
Among them, Song \etal \cite{song2019polysemous} used a biGRU to extract global features and a multi-head self attention to extract local features. Different form~\cite{song2019polysemous} that only used one video feature, Wang \etal~\cite{wang2021t2vlad} adopted seven different video features. They proposed a model named Text-to-Video VLAD (T2VLAD)~\cite{wang2021t2vlad}, where the video encoder proposed in \cite{liu2019use} was employed as the global branch and a multi-head self attention with a NetVLAD layer~\cite{arandjelovic2016netvlad} was used as the local branch.

Our proposed method is also multi-branch. Indeed, the idea of multiple-branch learning is not new by itself. In the context of video annotation, Carreira \etal \cite{carreira2017quo} utilized an RGB branch and an optical-flow branch for action recognition, and Liu \etal \cite{liu2020sibnet} used a content branch to encode salient visual content and a semantic branch to 
capture the semantic information in videos for video captioning.
However, different from the previous methods where two branches are independent, our design has two dependent branches. In particular, the proposed intensive-reading branch is aware of the representation learned by the other previewing branch. Such a novel design helps learn better video representation than the prior art.
The SlowFast network~\cite{feichtenhofer2019slowfast} also uses a multi-branch architecture, wherein the output of one branch depends on the other branch. Our proposed network differs from SlowFast in terms of the network input and how the dependency is exploited. Concerning the input, SlowFast feeds its two branches with video frames in different sampling rates and resolutions, while our model takes a sequence of frame-level features as a single input. The input is fed into two distinct encoders to derive complementary video representations. Concerning the dependency modeling, SlowFast makes its two branches dependent with lateral connections implemented by simple summation or concatenation, while we devise a previewing-aware attention (PaA) to connect the two branches explicitly. PaA is found to be better than simple concatenation, see Table~\ref{tab:ablation-fusion}.

\begin{figure*}[tb!]
\centering\includegraphics[width=2\columnwidth]{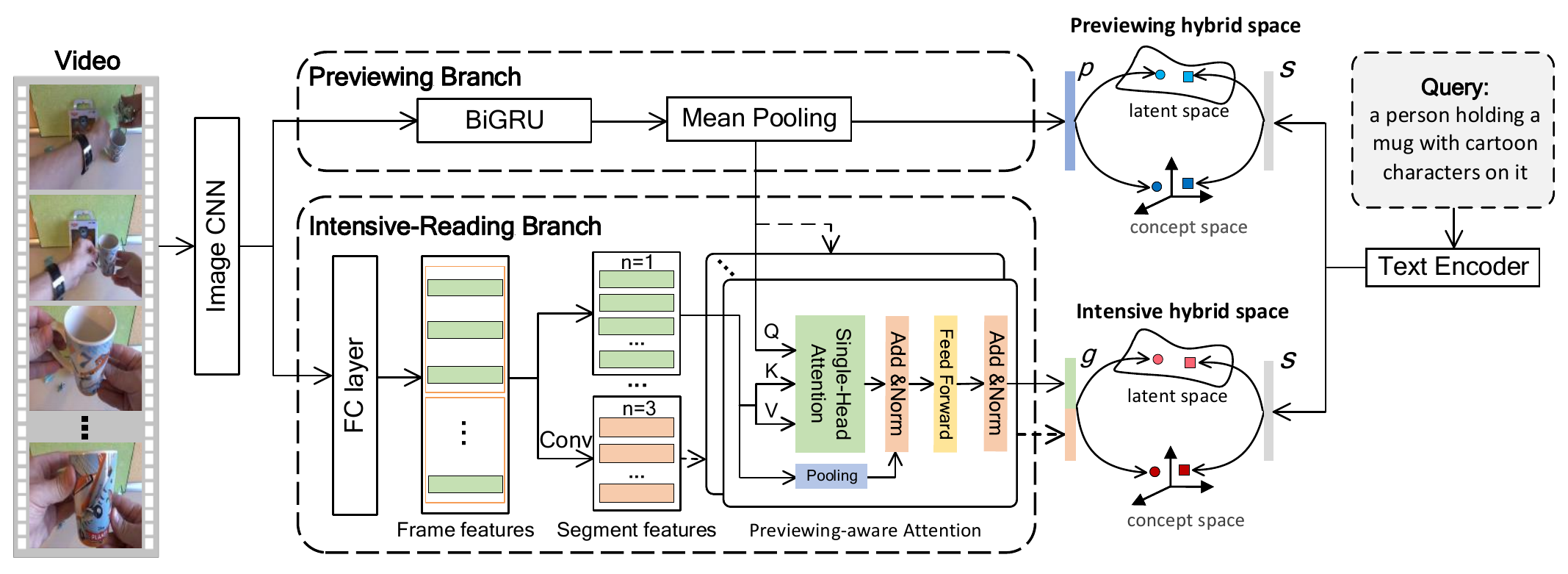}
\caption{\textbf{The proposed visual representation learning model for text-to-video retrieval}. Given an input video initially represented by a sequence of frame-level features, our model performs visual representation learning through two branches, \ie a previewing branch and an intensive-reading branch. The previewing branch is designed to capture the holistic information of the video, while the intensive-reading branch is to obtain more in-depth information.
The two branches are connected via Previewing-aware Attention (PaA) which makes the intensive-reading branch extract feature under the guidance of the previewing branch.
For query representation, we simply use off-the-shelf text encoders. }\label{fig:framework}
\vspace{-2mm}
\end{figure*}

\subsection{Cross-modal Similarity Computation}

Cross-modal similarity computation is one of the essential modules for text-to-video retrieval.
Popular solutions typically mapped videos and sentences into a common latent space, and the cross-modal similarities were measured by their distances in the latent space~\cite{song2021spatial, miech2019howto100m, dong2018predicting, dong2021multi}.
Additionally, there are some works~\cite{zhu2020actbert, yu2018joint,lei2021less,radford2021learning} that tend to directly predict cross-modal similarity without explicitly learning any common spaces. These works mainly focus on how to devise a cross-modal fusion module. 
As the representative work, Lei \etal \cite{lei2021less} concatenated frame sequence and word sequence, and fed them into a 12-layer transformer for cross-modal fusion. Besides, there is an increasing use of Transformer for multimodal fusion in the context of various tasks~\cite{yu2019deep, liu2021multimodal, hu2020iterative, yu2019multimodal}. For example, in the task of visual question answering, Yu \etal \cite{yu2019deep} proposed a guided attention module to fuse image and textual query features, and the attention modules can be cascaded in depth. In \cite{yu2019multimodal}, a multimodal Transformer model was extended for image captioning, where complex multimodal reasoning was conducted via multimodal Transformer. The above Transformer-based multimodal fusion can also be extended for text-to-video retrieval. Although using an advanced cross-modal fusion is likely to achieve performance gain for text-to-video retrieval, it is at the expense of efficiency.
As we focus on video representation learning, in this work we directly utilize the hybrid space proposed in~\cite{dong2021dual} for cross-modal similarity computation.

\section{Reading-strategy Inspired Visual Representation} \label{sec:method}

To represent a video containing complex events, we propose a reading-strategy inspired visual representation learning model.
As illustrated in Fig. \ref{fig:framework}, the proposed representation model consists of two branches: a previewing branch and an intensive-reading branch. 
The previewing branch is designed to capture coarse overview information that can be obtained by a lightweight module. The intensive-reading branch is designed to obtain more in-depth information, which can be captured by a relatively stronger module with the guidance of the overview information obtained by the previewing branch.
Note that the two branches are not independent. The intensive-reading branch is aware of the representation learned by the previewing branch, hence it can adaptively extract essential information according to the video overview. In what follows, we first depict the frame-level feature extraction, followed by the description of two branches respectively.

\subsection{Frame-Level Feature Extraction}
Before feeding videos into the previewing branch and the intensive-reading branch, videos are firstly encoded into a sequence of frame-level feature vectors, denoted as $V = \{v_1,v_2,\ldots,v_m\}$, where $v_i$ indicates the feature vector of the $i$-th frame and $m$ is the total number of video frames.
Following the common practice~\cite{dong2016early,li2020sea}, we extract the frame-level features by a pre-trained 2D CNN. 
Specifically, given a video, we first uniformly extract a sequence of video frames with a pre-specified interval of 0.5 seconds, and employ a 2D CNN pre-trained on ImageNet to extract the frame features.

\subsection{Previewing Branch}
For the previewing branch, we adopt bidirectional Gated Recurrent Unit (biGRU)~\cite{cho2014learning} to extract the overview information of the given video, considering it effective for sequence representation~\cite{dong2019dual,sigir2020tree}. BiGRU consists of a forward GRU and a backward GRU, where the forward GRU encodes frame features in normal order, while the backward one encodes them in reverse order. 
We feed the feature sequence $V$ to the biGRU.
The hidden states of both forward one and backward one are concatenated as the output of biGRU at a specific time step of $t=1,\ldots,m$, denoted as $h^t$.
Putting the outputs of all the time steps, we obtain a sequence of feature vectors $ H=\{h^1, h^2,..., h^m\}$ carried the whole information of the video.
Finally, through the previewing branch, the representation of the video $v$ is obtained by applying mean pooling on $H$ along the temporal dimension, which is denoted as $p \in \mathbb{R}^{1 \times d}$:
\begin{equation}
p = \frac{1}{m}\sum_{t=1}^m h^t,
\end{equation}
where $d$ indicates the dimensionality of the output vector.
Note that the previewing branch can be any video representation model that works on a video as a sequence of feature vectors.
Besides, the motivation behind our previewing branch is similar to \cite{lei2021less, wu2019liteeval, wu2019adaframe} which also use a lightweight branch to obtain the video overview. But these works adopt different strategies to improve the efficiency of video understanding models, while we aim for a better video representation.

\subsection{Intensive-Reading Branch}

As videos typically contain multiple objects and complex scenes, the previewing branch alone is insufficient to capture the information comprehensively. 
We introduce an intensive-reading branch that extracts more in-depth information with a strong structure under the guidance of the representation learned by the previewing branch.
The intensive-reading branch first represents video as multi-granularity segments, and then aggregates them with previewing-aware attentions.

\textbf{Multi-granularity Segment Representation}.
Inspired by the n-gram language model that considers a contiguous sequence of $n$ words from a given sentence to predict the probability of an entire sentence, we consider a sequence of $n$ contiguous frames from a given video to obtain more fine-grained video representation. We regard $n$ contiguous frames as a video segment.
Given a video, we split the video into a sequence of segments by a sliding window of a specific size, and represent segments of the same length as a sequence of feature vectors. 
Using multiple sliding windows of various sizes, we can obtain multiple sequences of segment feature vectors that are regarded as the multi-granularity segment representation.

More formally, given a video, we first employ an FC layer to map the frame features to a lower dimensional space, which is helpful for reducing the complexity of the model. After the feature mapping, we obtain a sequence of new mapped feature vectors $V^{'} = \{v_1^{'},v_2^{'},\ldots,v_m^{'}\}$. 
We then employ CNNs to obtain segment features, considering its good nature of aggregating adjacent features. Specifically, we adopt 1D convolutional layer over the frame features to obtain the segment features fusing the information of the adjacent frames. 
Let $Conv1D_{r,n,s}$ be a 1D convolutional layer that contains $r$ filters with size $n$ and stride $s$ to aggregate $n$ adjacent frames. 
The filter size  $n$ is the size of the sliding window. Setting $n$ to $1$ means each frame is treated as a segment, while using a larger $n$ would generate segments of a larger length.
The stride $s$ is the distance between adjacent segments. Smaller $s$ leads to more segments.
Given a sequence of new mapped feature vectors~$V^{'}$, the representation of segments generated with a sliding window of size $n$ is expressed as $C^{\, n} = \delta(Conv1D_{r,n,s}(V^{'}))$,
where $\delta$ indicates the ReLU activation.
Putting the segment representations generated with sliding windows of different sizes together, we obtain the multi-granularity segment representation as:
\begin{equation} \label{eq:seg-feat}
C = \{C^{\, n} \in \mathbb{R}^{ m_n \times r}   \}_{n\in \Phi },
\end{equation} 
where $\Phi$ denotes the set of sliding window sizes, $m_n$ indicates the number of segments of length $n$, and $r$ is the dimensionality of segment features. Note that $Conv1D$ with varied filter sizes are executed in parallel in one forward pass, the computation of Eq. \ref{eq:seg-feat} is efficient.

\textbf{Previewing-aware Attention}.
Given the multi-granularity segment representation $C$, we would like to enhance segments which are more important for conveying the semantic of videos. To this end, we propose a previewing-aware attention that adaptively selects video segments. 
We perform the previewing-aware attention for each granularity in parallel, and concatenate the output as the final output of the intensive-reading branch.

\begin{figure}[tb!]
\centering
\subfigure[Self Attention]{
\includegraphics[width=0.9\columnwidth]{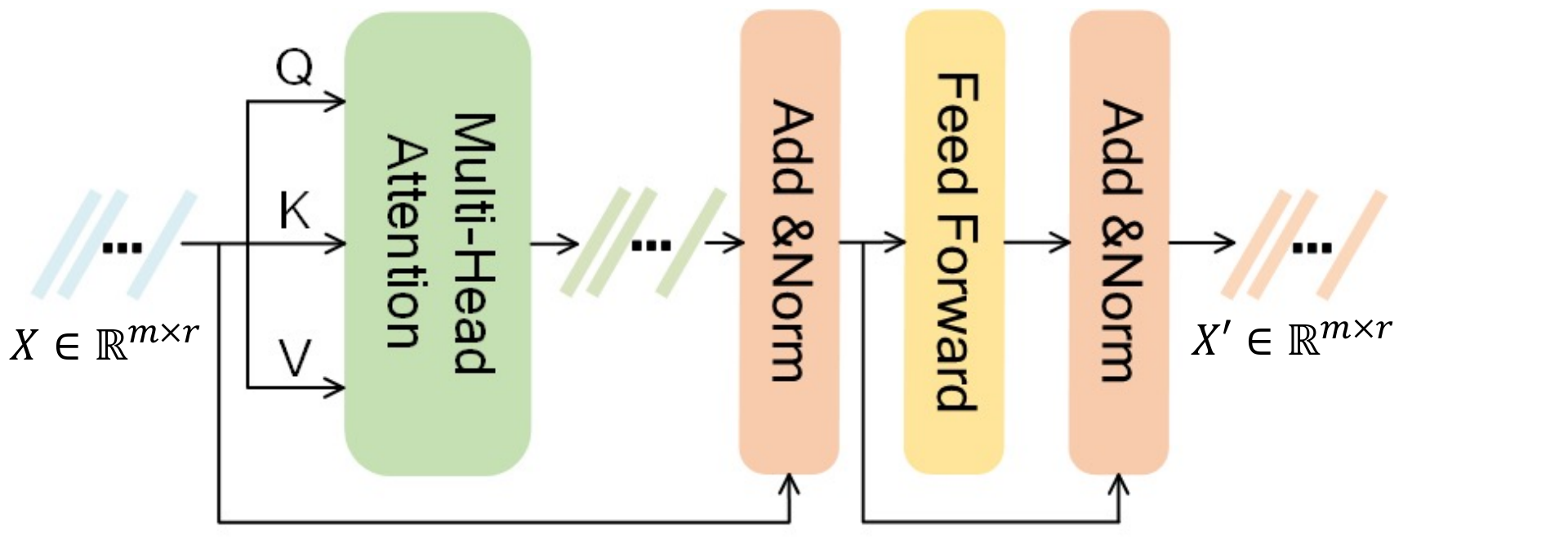}
}
\quad
\subfigure[Previewing-aware Attention]{
\includegraphics[width=0.9\columnwidth]{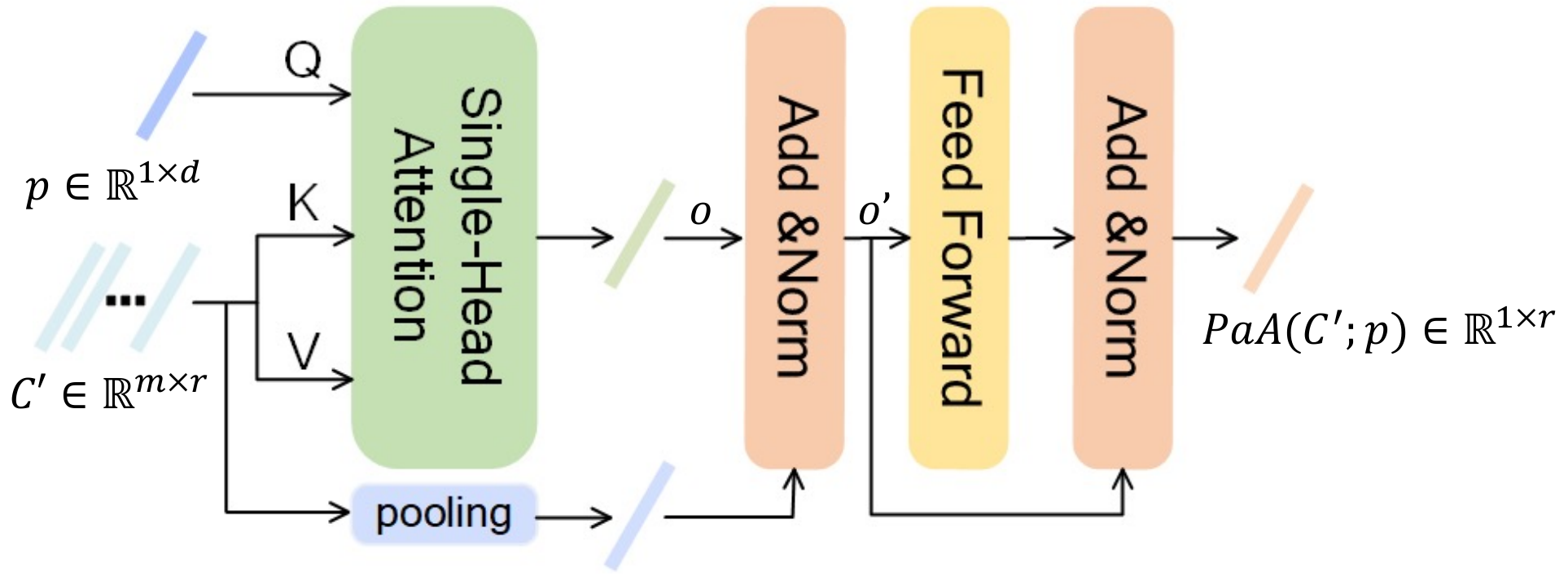}
}
\caption{ \textbf{Attention illustration}. (a) Self attention takes the segment features as the input and maps them into queries, keys, values.
By contrast, (b) previewing-aware attention uses the previewing feature vector as the query, and the segment features as keys and values, which adaptively selects segments conveying the semantic of the video under the guidance of the previewing feature.
}\label{fig:attention_pic}
\end{figure}

To devise the previewing-aware attention, we borrow the idea of multi-head self-attention (MHSA) mechanism in Transformer~\cite{vaswani2017attention}, considering its decent performance as an attention module in a variety of computer vision related tasks~\cite{chen2020image,zhong2020self}.
MHSA first projects the input into queries, keys, values, and then computes the output as a weighted sum of the values. The weight assigned to each value is computed by a compatibility function of the query with the corresponding key.
We adapt MHSA by taking the previewing feature vector $p$ as the query, and the segment features (of a specific granularity)  $C'$ as keys and values, see Fig.~\ref{fig:attention_pic}.
Specifically,  $p$ is first linearly projected into $d_k$-dimensional vector as the query, and $C'$ are linearly projected into $d_k$-dimensional and $d_v$-dimensional vectors used as the keys and values respectively. Then a scaled dot-product attention with an extra output project layer is employed. An attended output $o$  is obtained as:
\begin{equation}
\left\{\begin{array}{l}
 Attention(Q, K, V) := Softmax(\frac{QK^T}{\sqrt{d_k}})V,  \\
o = W_4 \times Attention( p W_1, C' W_2, C' W_3 ), \\
\end{array} \right.
\end{equation}
where $Attention$ denotes the scaled dot-product attention, $W_1 \in \mathbb{R}^{d\times d_k}$, $W_2 \in \mathbb{R}^{r\times d_k}$, $W_3 \in \mathbb{R}^{r \times d_v}$ are three projection matrices for transforming the input features, and $W_4 \in \mathbb{R}^{d_v \times d_v}$ is the output projection matrix. Different from MHSA, we use one head only, as we found in our preliminary experiment that the gain of using more heads is marginal. We attribute this to the different contexts we are dealing with. Recall that the multi-head trick was originally used for mapping the input feature into different and hopefully complementary representation subspaces \cite{vaswani2017attention}. In our context, by multi-granularity segmentation representation learning, we have constructed multiple inputs, each fed into its own PaA block with the output concatenated later, see Fig. \ref{fig:framework}. Therefore, even with one-head PaA, we have practically implemented the notion of multi-subspace learning and fusion, and thus making multi-head unnecessary.

Following~\cite{vaswani2017attention}, we enhance $o$ with a residual connection. A max pooling operation is performed on the input $C'$ so it can be added to $o$. 
The enhanced output $o'$ is given by:
\begin{equation}
    o' =  LN(o + maxpool(C')),
\end{equation}
where $LN$ indicates the layer normalization.
Next, a feed-forward network with a residual connection and a layer normalization is employed. 
We implement the feed-forward network as a multi-layer perceptron (MLP) consisting of two FC layers with a ReLU activation.
The final output of the PaA module, guided by the previewing-branch output feature $p$, is obtained as
\begin{equation}
\begin{array}{cc}
    PaA(C'; p) = LN(o' + MLP(o')).   
\end{array}
\end{equation}
As illustrated in Fig. \ref{fig:attention_pic} (b), PaA transforms a given input $C'$ of size  $m \times r$ to an output of size $1 \times r$. Different from MHSA, PaA essentially performs one-round feature aggregation. Hence, stacking multiple blocks as MHSA uses is unsuited for PaA.

Given the multi-granularity segment representation, we perform previewing-aware attention for each granularity in parallel, and concatenate the output as the final output $g$ of the intensive-reading branch. Formally, the output $g$ is obtained as:
\begin{equation}
\begin{array}{cc}
g = \underset{n\in \Phi }{Concat}(PaA(C^n; p)),\\
\end{array}
\end{equation}
where $Concat$ indicates the concatenation operation.
Note that the intensive-reading branch allows the model to jointly extract information from segment representations of different granularities, thus obtaining more fine-grained video representation.

\section{Video-Text Matching}

\subsection{Text Representation}
As our focus is on the video side, we adopt multi-level text encoding~\cite{dong2021dual} and BERT~\cite{devlin2018bert} for text representation, considering their open-sourced code and good performance in the video-text matching task.
Given a sentence, the multi-level text encoding extracts three levels of textual features based on Bag-of-Words~(BoW), biGRU, biGRU-CNN, and concatenates them as its output feature vector.
For BERT, we utilize it trained on Wikipedia and book corpora~\cite{zhu2015aligning}, and generate a 1,024-dimensional feature vector for each sentence.
Specifically, we represent text in two ways, one is only using the multi-level text encoding, the other one is using the concatenation of both encoders' outputs as the text representation.
Multi-level text encoding is utilized as the default text representation unless otherwise stated.
For ease of reference, the representation of a sentence is denoted as $s$.

\subsection{Video-Text Similarity learning}
For video-text similarity learning, we employ hybrid space learning~\cite{dong2021dual} which combines the high performance of the latent space and the good interpretability of the concept space.
Different from previous work~\cite{dong2021dual} that learns one hybrid space, we use a double hybrid spaces learning strategy which learns two hybrid spaces, one for the previewing branch and the other for the intensive-reading branch. 
The final video-text similarity is the aggregation of the video-text similarities in the two hybrid spaces.

\textbf{Double Hybrid Spaces Learning.}
A hybrid space is comprised of a latent space and a concept space.
Given the encoded video feature vector $v'$ and the sentence feature vector $s'$, the hybrid space learning projects them into a latent space and a concept space via an FC layer respectively.
For the latent space learning, it is expected to make relevant video-sentence pairs near and irrelevant pairs far away in the space. Therefore, an improved triplet ranking loss \cite{fartash2017vse++} is employed, which are commonly used in retrieval-based tasks~\cite{li2020sea,yang2018person,liu2020deep,dong2021fine}.
For the concept space learning, it can be naturally formulated as a multi-label classification problem. 
As the concept space is expected to be used for both interpretability and video-text matching, a binary cross-entropy loss and an improved triplet ranking loss are jointly used for space learning. For ease of reference, we denote the joint loss of learning a hybrid space as $\mathcal{L}_{v',s'}$. We refer the readers to~\cite{dong2021dual} for more detailed descriptions.

In our model, we propose to learn two hybrid spaces, one for the previewing branch and the other for the intensive-reading branch. We name the former as the previewing hybrid space and the latter as the intensive hybrid space.
Specifically, given a video encoded as a previewing feature vector $p$ and a intensive-reading feature vector $g$, and a sentence represented as a feature vector $s$, we learn the previewing hybrid space based on $p$ and $s$ and the intensive hybrid space based on $g$ and $s$, respectively. The whole model is trained via minimizing the sum of previewing hybrid space loss $\mathcal{L}_{p,s}$ and the intensive hybrid space loss $\mathcal{L}_{g,s}$:
\begin{equation}\label{eq:loss}
\mathcal{L} =  \mathcal{L}_{p,s} + \mathcal{L}_{g,s}.
\end{equation}

\textbf{Video-Text Similarity.}
Once the model is trained, the final similarity between a video $v$ and a sentence $s$ is computed as the sum of  similarities in the two hybrid spaces, that is:
\begin{equation}\label{eq:final_simi}
sim(v,s) =  sim_{p}(v,s) + sim_{g}(v,s),
\end{equation}
where $sim_{p}(v,s)$ and $sim_{g}(v,s)$ indicates the video-text similarity in the previewing hybrid space and the intensive hybrid space, respectively.
Detailed calculations of $sim_{p}(v,s)$ and $sim_{g}(v,s)$ can be found in the appendix.

\section{Experiments} \label{sec:eval}

\subsection{Experimental Settings} \label{ssec:exp-set}
\subsubsection{Datasets}
We use three public datasets: MSR-VTT~\cite{xu2016msr}, VATEX \cite{wang2019vatex} and TGIF \cite{tgif}. Table \ref{tab:msrvtt_dataset} summarizes the brief statistics of these datasets.

\textit{MSR-VTT} \cite{xu2016msr}: 
MSR-VTT is a commonly used dataset for text-to-video retrieval task. This dataset contains 10k web videos, and each video is associated with 20 crowd-sourced natural language sentences. The sentences briefly describe the content of the corresponding video. 
Following the recent work~\cite{dong2021dual},  we conduct experiments on three distinct editions of data split~\cite{yu2018joint,miech2018learning,xu2016msr}, which we refer to as MV-Yu, MV-Miech and MV-Xu, respectively.

\textit{TGIF} \cite{tgif}: 
TGIF is originally developed for the captioning task, while now also popular for the text-to-video retrieval task. It contains 100K animated GIFs collected from Tumblr, and 120K natural language sentences annotated via crowdsourcing. Each GIF is associated with 1-3 sentences. 
For this dataset, we notice there are two distinct editions of data split in the literature \cite{chen2020fine,li2020sea}.
For ease of reference, we term the two splits as TGIF-Chen~\cite{chen2020fine} and TGIF-Li~\cite{li2020sea}, respectively.
We conduct experiments on all the two data splits for a comprehensive evaluation.

\textit{VATEX}~\cite{wang2019vatex}:
VATEX is a large-scale multilingual video description dataset. 
Each video, collected for YouTube, has a duration of about 10 seconds. 
Each video has corresponding 10 English sentences and 10 Chinese sentences to describe the video content. We only use the English sentences in our experiments. 
Following \cite{dong2021dual, chen2020fine}, we utilize 25,991 videos for training, 1,500 videos for validation and 1,500 videos for testing. Note that the validation and test set are obtained by randomly splitting the official validation set of 3,000 videos into two equal parts.

\subsubsection{Implementation Details}
We use PyTorch as our deep learning environment. 
For sentence preprocessing, following~\cite{dong2021dual} we first convert all words to the lowercase and then replace words that are occurring less than five times in the training set with a special token.
For video features, on MSR-VTT, we use the frame-level ResNeXt-101~\cite{xie2017aggregated,mettes2020shuffled} and ResNet-152~\cite{cvpr2016-resnet} provided in \cite{dong2021dual}. The two feature vectors are concatenated to obtain a combined 4,096-d CNN feature, which we refer to as ResNeXt-ResNet.
On TGIF, for a fair comparison, we utilize frame-level ResNet-152 feature provided by~\cite{chen2020fine} on the TGIF-Chen split, and ResNeXt-ResNet provided by~\cite{li2020sea} on the TGIF-Li split.
On VATEX, we adopt 1,024-d I3D \cite{carreira2017quo} video features provided by the dataset developers~\cite{wang2019vatex}.

For model structure, in the previewing branch, we let the hidden size of GRU be 512, so the dimensionality $d$ of the output feature is 1,024.
In the intensive-reading branch, we set the dimensionality of the mapped frame feature to 2048, and let the filter number $r$ be 1024 and the stride size $s$ be 2 in all convolutional layers.
$d_k$ and $d_v$ in the attention are set to 512, and 1024, respectively.
The default set of sliding window sizes is $\Phi$=$\{1,3\}$ unless otherwise stated.
For the multi-level text encoding and hybrid space learning implementation, we directly follow the source code\footnote{https://github.com/danieljf24/hybrid\_space} provided in~\cite{dong2021dual}.

For model training, we utilize an Adam optimizer with a mini-batch size of 128.
The initial learning rate is set to 0.0001.
We take an adjustment schedule similar to \cite{dong2018predicting,dong2021dual}. That is, once the validation loss does not decrease in three consecutive epochs, we divide the learning rate by 2. Early stop occurs if the validation performance does not improve in ten consecutive epochs. The maximal number of epochs is 50. Additionally, for the ease of model training, we fix the parameters of the BERT if it is used.

\begin{table} [tb!]
\renewcommand{\arraystretch}{1.2}
\caption{Brief statistics of three public datasets used in our experiments: MSR-VTT, TGIF and VATEX. Note that for MSR-VTT and TGIF, multiple data splits exist. We include them for a fair and comprehensive comparison.}
\label{tab:msrvtt_dataset}
\centering 
\scalebox{0.85}{
\begin{tabular}{l*{7}{r}}
\toprule
\multirow{2}{*}{\textbf{DataSets}}   & \multicolumn{3}{c}{\textbf{\#Videos / GIFs}} && \multicolumn{3}{c}{\textbf{\#Sentences}}  \\
 \cmidrule{2-4} \cmidrule{6-8}
& train & val & test  && train & val & test \\
\cmidrule{1-8}
\textbf{MSR-VTT}~\cite{xu2016msr}  & & & & & \\
MV-Yu~\cite{yu2018joint}                   & 7,010 & 1,000 & 1,000   && 140,200  & 1,000 & 1,000\\
MV-Miech~\cite{miech2018learning}          & 6,656  & 1,000 & 1,000  && 133,120  & 1,000 & 1,000 \\
MV-Xu~\cite{xu2016msr}                    & 6,513  & 497 & 2,990   && 130,260  & 9,940 & 59,800  \\ [2pt]
\hline
\textbf{TGIF} \cite{tgif} & & & & & \\
TGIF-Chen~\cite{chen2020fine}    & 79,451  & 10,651 & 11,310   && 80,295  & 10,774 & 33,951  \\
TGIF-Li~\cite{li2020sea}    & 78,799  & 10,705 & 11,351   && 79,632  & 10,828 & 34,074  \\ [2pt]
\hline
\textbf{VATEX}~\cite{wang2019vatex}                    & 25,991  & 1,500 & 1,500   && 259,910  & 15,000 & 15,000  \\
\bottomrule
\end{tabular}
 }
\end{table}

\subsubsection{Evaluation Metrics}
Following the previous works~\cite{li2019w2vv++,dong2021dual}, we use rank-based metrics, namely $R@K$ ($K = 1, 5, 10$), Median rank (Med r) and mean Average Precision (mAP) to evaluate the performance. $R@K$ is the fraction of queries that correctly retrieve desired items in the top $K$ of the ranking list. Med r is the median rank of the first relevant item in the search results. Higher $R@K$, mAP and lower Med r mean better performance. 
For overall comparison, we report the Sum of all Recalls (SumR).

\subsection{Comparison with the State-of-the-art} \label{ssec:exp-sota} 

\begin{table} [tb!]
\renewcommand{\arraystretch}{1.2}
\caption{Performance comparison on the MV-Yu split of MSR-VTT. Symbol asterisk (*) indicates that models utilize the same video feature. The best and second-best results are in \textbf{bold} and \underline{underlined}, respectively. 
"-" indicates the corresponding result is unavailable from the original papers.
Our proposed model using two features compares favorably to recent state-of-the-art model T2VLAD using seven features.}
\label{tab:sota-msrvtt}
\centering 

\scalebox{0.8}{
	\begin{tabular}{lc*{6}{r}c @{}}
	\toprule  
 
  \textbf{Method}  & \textbf{BERT?} & \textbf{R@1} & \textbf{R@5} & \textbf{R@10} & \textbf{Med r} & \textbf{mAP} && \textbf{SumR} \\
    \cmidrule{1-9}
CT-SAN   \cite{yu2017end}                 & & 4.4 & 16.6 & 22.3 & 35 & - &&  43.3  \\
JSFusion \cite{yu2018joint}              &  & 10.2 & 31.2 & 43.2 & 13 & - &&  84.6  \\
STG \cite{song2021spatial}               &  & 15.5 & 39.2 & 50.4 & 10 & - &&  105.1 \\
TCE \cite{sigir2020tree}                 &  & 16.1 & 38.0 & 51.5 & 10 & - &&  105.6  \\
Miech \etal  \cite{miech2019howto100m}   &  & 14.9 & 40.2 & 52.8 & 9 & - &&  107.9 \\
UniVL \cite{luo2020univl}                & \checkmark  & 16.7 & 44.0 & 55.9 & 6 & - &&  116.6 \\
CE \cite{liu2019use}                    & \checkmark  & 20.9  & 48.8 & 62.4  & 6 & - && 132.1 \\
MMT \cite{gabeur2020multi}              & \checkmark & 24.6 & 54.0 & 67.1 & \uline{4} & - &&  145.7  \\
Support-Set \cite{patrick2020support}   & \checkmark  & 27.4 & 56.3 & 67.7 & \textbf{3} & - && 151.3\\
CMGSD \cite{he2021improving}            & \checkmark & 26.1 & 56.7 & 69.7 & \uline{4} & - &&  152.5  \\
T2VLAD \cite{wang2021t2vlad}   & \checkmark  & \uline{29.5} & \uline{59.0} & \uline{70.1} & \uline{4} & - && \textbf{158.6}\\

\cmidrule{3-9}
W2VV* \cite{dong2018predicting}          &  & 1.9 & 9.9 & 15.2 & 79 & 6.8 &&   27 \\
VSE++* \cite{fartash2017vse++}           &    & 16.0 & 38.5 & 50.9 & 10 & 27.4 &&  105.4 \\
MEE* \cite{miech2018learning}           &  & 14.6 & 38.4 & 52.4 & 9 & 26.1 &&  105.4 \\
W2VV++* \cite{li2019w2vv++}             &   & 19.0 & 45.0 & 58.7 & 7 & 31.8 &&  122.7 \\
CE* \cite{liu2019use}                  & \checkmark  & 17.2	& 46.2 & 58.5 & 7  & 30.3 &&  121.9 \\
TCE* \cite{sigir2020tree}              &   & 17.8 & 46.0 & 58.3 & 7 & 31.1 &&  122.1 \\
HGR* \cite{chen2020fine}               &   & 21.7 & 47.4 & 61.1 & 6 & 34.0 &&  130.2 \\
Dual Encoding* \cite{dong2021dual}     & & 21.1 & 48.7 & 60.2 & 6 & 33.6 &&  130.0 \\
SEA* \cite{li2020sea}       &  \checkmark  & 23.8 & 50.3 & 63.8 & 5 & 36.6 &&  137.9 \\
RIVRL*                 &  &23.3 & 52.2 & 63.8 & 5 & 36.7 &&  139.3 \\
RIVRL with BERT*       & \checkmark & 27.9 & \textbf{59.3} & \textbf{71.2} & \uline{4} & \textbf{42.0} &&  \uline{158.4}   \\

\cmidrule{1-9}
\multicolumn{5}{l}{\textbf{Pre-training on HowTo100M~\cite{miech2019howto100m}:}} \\
UniVL \cite{luo2020univl} &\checkmark  &20.6 & 49.1 & 62.9 & 6 & - && 132.6\\
AVLnet \cite{rouditchenko2020avlnet}&  & 22.5 & 50.5 & 64.1 & 5 &- && 137.1\\
MMT \cite{gabeur2020multi}          & \checkmark  & 26.6 & 57.1 & 69.6 & \uline{4} & - &&  153.3  \\
Support-Set \cite{patrick2020support}   &\checkmark  & \textbf{30.1} & 58.5 & 69.3 & \textbf{3} & - && 157.9 \\
\cmidrule{1-9}
\multicolumn{5}{l}{\textbf{Pre-training on TV~\cite{lei2018tvqa}:}} \\
HERO \cite{li2020hero} &\checkmark  & 16.5  & 40.0  & 52.0  & -   & -  && 108.5   \\
ActBERT \cite{zhu2020actbert} &\checkmark  & 11.5 & 34.0  & 47.3  & -   & -  && 92.8   \\
CoCo-BERT \cite{luo2021coco} & \checkmark   & 18.6 & 43.8 & 56.2  & -   & -  && 118.6   \\
\cmidrule{1-9}
\multicolumn{5}{l}{\textbf{Pre-training on ACTION~\cite{pan2020auto}:}} \\

HERO \cite{li2020hero} &\checkmark  & 17.7  & 41.2   & 53.0   & -   & -   && 111.9   \\
ActBERT \cite{zhu2020actbert} &\checkmark  & 13.2  & 37.2  & 50.4  & -   & - && 100.8  \\
CoCo-BERT \cite{luo2021coco} &\checkmark  & 21.1 & 45.8   & 57.2   & - & - && 124.1  \\

    \bottomrule
    \end{tabular}
 }
\end{table}

\begin{table} [!tb]
\renewcommand{\arraystretch}{1.2}
\caption{Performance comparison on the MV-Miech and MV-Xu splits of MSR-VTT. Symbol asterisk (*) indicates that models utilize the same video feature. Our proposed model compares favorably to state-of-the-art models.
}
\label{tab:sota-msrvtt-others}
\centering 

\scalebox{0.8}{
	\begin{tabular}{@{}lc*{6}{r}c @{}}
	\toprule  
  \textbf{Method}  & \textbf{BERT?} & \textbf{R@1} & \textbf{R@5} & \textbf{R@10} & \textbf{Med r} & \textbf{mAP} && \textbf{SumR} \\
    \cmidrule{1-9}
    
\textbf{MV-Miech \cite{miech2018learning}}  \\
JPoSE  \cite{wray2019fine}                & & 14.3 & 38.1 & 53.0 & 9 & - &&    105.4  \\
MEE \cite{miech2018learning}               &    & 16.8 & 41.0 & 54.4 & 9 & - &&    112.2   \\
TCE \cite{sigir2020tree}                &  & 17.1 & 39.9 & 53.7 & 9 & - &&  110.7  \\
CE  \cite{liu2019use}                    &\checkmark  & 18.2 & 46.0 & 60.7 & 7 & - &&  124.9 \\
MMT \cite{gabeur2020multi}     &\checkmark   & 20.3 & 49.1 & 63.9 & 6 & - &&  133.3  \\
CMGSD \cite{he2021improving}    &\checkmark   & 22.7 & 52.6 & 66.1 & 6 & - &&  141.4  \\
T2VLAD \cite{wang2021t2vlad}   &\checkmark    & \uline{26.1} & \uline{54.7} & \textbf{68.1} & \textbf{4}  & - &&  \textbf{148.9}  \\

\cmidrule{3-9}
W2VV* \cite{dong2018predicting}            &  & 2.7 & 12.5 & 17.3 & 83 & 7.9 &&   32.5 \\
MEE* \cite{miech2018learning}              &  & 15.7 & 39.0 & 52.3 & 9 & 27.1  &&  107 \\
VSE++* \cite{fartash2017vse++}              &    & 17.0 & 40.9 & 52.0 & 10 & 16.9 &&  109.9 \\
CE* \cite{liu2019use}                     &   & 17.8 & 42.8 & 56.1 & 8 & 30.3  &&  116.7 \\
TCE* \cite{sigir2020tree}                &    & 17.0 & 44.7 & 58.3 & 7 & 30.0  &&  120 \\
W2VV++* \cite{li2019w2vv++}              &    & 21.7 & 48.6 & 60.9 & 6 & 34.4 &&   131.2 \\
HGR* \cite{chen2020fine}                 &    & 22.9 & 50.2 & 63.6 & 5 & 35.9  &&  136.7 \\
Dual Encoding* \cite{dong2021dual}       &    & 23.0 & 50.6 & 62.5 & 5 & 36.1 &&  136.1 \\
RIVRL*                &   & 25.8 & 53.7 & 67.0 & 5 & \uline{38.5} && 146.5 \\
RIVRL with BERT*     &\checkmark    & \textbf{26.2} & \textbf{55.4} & \uline{67.3} & \textbf{4} & \textbf{39.6} &&  \textbf{148.9}  \\
[3pt]

\hline
\textbf{MV-Xu \cite{xu2016msr}}  \\
Francis \etal \cite{iccv2019-francis}     && 6.5 & 19.3 & 28.0 & 42 & - && 53.8  \\
Mithun \etal \cite{mithun2018learning}    && 7.0 & 20.9 & 29.7 & 38 & - &&  57.6 \\
TCE \cite{sigir2020tree}                  && 7.7 & 22.5 & 32.1 & 30 & - &&  62.3  \\
ViSERN \cite{feng2020exploiting}          && 7.9 & 23.0 & 32.6 & 30 & - && 63.5  \\
CF-GNN \cite{wang2020learning}            && 8.0 & 23.2 & 32.6 & 31 & 16.0  && 63.8  \\
CVTR \cite{li2020novel}             && 7.8 & 23.2 & 33.5 & 28 & - &&  64.5  \\
STG \cite{song2021spatial}          && 8.3  & 23.7 & 33.9 & 28 & - &&  65.9 \\
MSDC \cite{zhao2020stacked}     &\checkmark & 8.8 & 25.5 & 36.5 & 22 & 17.4 && 70.8  \\
GPO \cite{chen2021learning}   && 9.1  & 25.9 & 36.3 & - & - &&  71.3   \\
HGR \cite{chen2020fine}            && 9.2  & 26.2 & 36.5 & 24 & - &&  71.9 \\
CE \cite{liu2019use}               &\checkmark & 10.0 & 29.0  & 41.2  & 16  & -   && 80.2 \\
CE+UWML \cite{wei2021universal}    &\checkmark & 10.9 & 30.4  & 42.3  & -  & -   && 83.6 \\
CE+IRA \cite{chen2020interclass}   &\checkmark & 11.2 & 31.1  & 42.8  & -  & 11.2   &&85.1 \\
CMGSD \cite{he2021improving}        &\checkmark & 11.3  & 32.0 & 44.1 & 14.2 & - &&  87.4 \\

T2VLAD \cite{wang2021t2vlad}    &\checkmark & 12.7 & \uline{34.8} & \textbf{47.1}  & \textbf{}{12}  & -   && \uline{94.6} \\
\cmidrule{3-9}
W2VV* \cite{dong2018predicting}    && 1.1  & 4.7  & 8.1 & 236 & 3.7 &&   13.9 \\
MEE* \cite{miech2018learning}      && 6.8 & 20.7 & 31.1 & 28 & 14.7  &&   58.6 \\
CE* \cite{liu2019use}              && 7.9 & 23.6 & 34.6 & 23 & 16.5 &&   66.1 \\
VSE++* \cite{fartash2017vse++}        && 8.7  & 24.3 & 34.1 & 28 & 16.9 &&   67.1   \\
TCE* \cite{sigir2020tree}          && 9.3 & 27.3 & 38.6 & 19 & 18.7 &&    75.2  \\
W2VV++* \cite{li2019w2vv++}        && 11.1 & 29.6 & 40.5 & 18 & 20.6 &&   81.2  \\
HGR* \cite{chen2020fine}           && 11.1 &	30.5 &	42.1 & 16 & 20.8 &&   83.7  \\
Dual Encoding*  \cite{dong2021dual}       && 11.6 & 30.3 & 41.3 & 17 & 21.2 &&    83.2  \\
SEA~*\cite{li2020sea} && 12.4 & 32.1 & 43.3 & 15 & 22.3 &&    87.8   \\
SEA with BERT~*\cite{li2020sea} &\checkmark & 13.1 & 33.4 & 45.0 & 14 & \uline{23.3} &&   91.5   \\
RIVRL* && \uline{13.0} & 33.4 & 44.8 & 14 & 23.2 && 91.2 \\
RIVRL with BERT*  &\checkmark & \textbf{13.8} & \textbf{35.0} & \uline{46.5} & \uline{13} & \textbf{24.3}  && \textbf{95.2}  \\
[3pt]

    \bottomrule
    \end{tabular}
 }
\end{table}

\subsubsection{Experiments on MSR-VTT}
Table \ref{tab:sota-msrvtt} summarizes the performance comparison on the MV-Yu split of MSR-VTT, where symbol asterisk (*) indicates that models utilize the same ResNeXt-ResNet video feature.
Among the models using the same video feature, our proposed model RIVRL consistently outperforms the others with a clear margin.
Among them, Dual Encoding~\cite{dong2021dual} utilizes the same multi-level text encoding and hybrid space learning with RIVRL, while using different video representation learning methods. The better performance of our model over Dual Encoding demonstrates the effectiveness of our proposed reading-strategy inspired visual representation learning.
Additionally, integrating BERT in the text encoder, RIVRL with BERT achieves significant performance improvements. Besides, it compares favorably to the recent state-of-the-art model T2VLAD \cite{wang2021t2vlad} which utilizes seven video features. By contrast, our model only utilizes two visual features, and is much more lightweight than T2VLAD (Please see the complexity analysis in Section \ref{ssec:complexity}).

Comparing the models pre-trained on a large-scale dataset HowTo100M, our best model without pre-trianing on it performs the best in terms of SumR. 
Among the four models pre-trained on HowTo100M, Support-Set performs the best and its R@1 score is better than ours. However, when comparing with Support-Set without pre-trianing on HowTo100M, our model gives the better performance except in terms of MedR. 
Additionally, comparing the models pre-trained on TV dataset~\cite{lei2018tvqa} or a large-scale GIF video dataset ACTION~\cite{pan2020auto}, our model still performs better.
The results show the effectiveness of our model for text-to-video retrieval.

Performance comparison on the MV-Miech and MV-Xu splits of MSR-VTT are shown in Table \ref{tab:sota-msrvtt-others}. 
Similarly on MV-Yu, our proposed model performs the best among the models using the same video feature, and compares favorably to state-of-the-art model T2VLAD that utilizes seven video features.
It is worth noting that, on MV-Xu, our model without BERT gives comparable performance to SEA~\cite{li2020sea} with BERT. 
SEA employs the simple mean pooling with three independent FC layers as the video encoder, while our model utilizes two dependent branches to represent videos. The result demonstrates the effectiveness of multiple dependent branches for video representation.

\begin{table} [tb!]
\renewcommand{\arraystretch}{1.2}
\caption{State-of-the-art on TGIF. On each split, all models utilize the same video feature. Our proposed model consistently performs the best.}
\label{tab:tgif_perf}
\centering 
\scalebox{0.9}{
\begin{tabular}{l*{7}{r}}
\toprule
  \textbf{Method}  & \textbf{R@1} & \textbf{R@5} & \textbf{R@10} & \textbf{Med r} & \textbf{mAP} && \textbf{SumR} \\
\hline
\textbf{TGIF-Chen}~\cite{chen2020fine}  \\
VSE++ \cite{fartash2017vse++}             & 0.4 & 1.6 & 3.6 &692 & - && 5.6  \\
Order \cite{2015Order}             & 0.5 & 2.1 & 3.8 &500 & - && 6.4 \\
Corr-AE \cite{feng2014cross}         & 0.9 & 3.4 & 5.6 &365 & - && 9.9 \\
DeViSE \cite{frome2013devise}     & 0.8 & 3.5 & 6.0 &378 & - && 10.3 \\
PVSE \cite{song2019polysemous}               & 2.3 & 7.5 & 11.9  &162 & - && 21.7 \\
HGR \cite{chen2020fine}                 & 4.5 & 12.4 & 17.8  &160 & - && 34.7 \\
HGR+UWML \cite{wei2021universal}        & 6.4 & 13.4   & 16.9   & - & - && 36.7 \\
RIVRL                                      & \uline{6.5} & \uline{16.2} & \uline{22.5} & \uline{92} & \uline{11.94} && \uline{45.2} \\
RIVRL with BERT                            & \textbf{6.9} & \textbf{16.8} & \textbf{23.4} & \textbf{82} & \textbf{12.43}  && \textbf{47.1} \\
\hline
\textbf{TGIF-Li}~\cite{li2019w2vv++} \\
PVSE \cite{song2019polysemous}                & 3.0 &9.7 &14.9 &109 & - && 27.6 \\
W2VV++ \cite{li2019w2vv++}                & 9.4 &22.3 &29.8 &48 &16.20 && 61.5 \\
CF-GNN \cite{wang2020learning}            & 10.2 &23.0 &30.7 &44 & - && 63.9 \\
SEA \cite{li2020sea}                      & 10.2 & 23.6 & 31.3 & 41 & 17.20 && 65.1 \\
Dual Encoding\cite{dong2021dual}          & 10.6 &23.7 &31.8 &39 &17.62 && 66.1 \\
SEA with BERT \cite{li2020sea}            & 11.1 & 25.2 & 32.8 & 35 & 18.50 && 69.1 \\ [2pt]
RIVRL                                      & \uline{11.3} & \uline{25.3} & \uline{33.5} & \uline{33} & \uline{18.75} && \uline{70.1} \\
RIVRL with BERT                            & \textbf{12.2} & \textbf{26.8} & \textbf{35.4} & \textbf{30} & \textbf{19.87} && \textbf{74.4} \\
\bottomrule
\end{tabular}
 }
\end{table}

\subsubsection{Experiments on TGIF}
In Table \ref{tab:tgif_perf}, we report performance on two distinct data splits of TGIF. 
On each split, all models utilize the same video feature. Our proposed RIVRL without BERT consistently outperforms others, including recent model SEA with BERT.
Moreover, our model achieves consistent performance improvement if we integrate BERT into the text encoder.
The results not only show the effectiveness of our proposed reading-strategy inspired visual representation for text-to-video retrieval, but also demonstrate its potential of using it with a stronger text encoder.

\begin{table} [tb!]
\renewcommand{\arraystretch}{1.2}
\caption{State-of-the-art on VATEX. All models are trained from scratch. The same I3D feature is used. Our proposed model performs the best.}
\label{tab:vatex_perf}
\centering 

\scalebox{0.92}{
	\begin{tabular}{@{}l*{6}{r}c @{}}
	\toprule  
\textbf{Method} &  \textbf{R@1} & \textbf{R@5} & \textbf{R@10} & \textbf{Med r} & \textbf{mAP} && \textbf{SumR}  \\
    \cmidrule{0-7} 
W2VV   \cite{dong2018predicting}     & 14.6 & 36.3 & 46.1 & - & -   &&   97.0 \\
VSE++  \cite{fartash2017vse++}       & 31.3 & 65.8 & 76.4 & - & -  &&   173.5   \\
CE     \cite{liu2019use}             & 31.1 & 68.7 & 80.2 & - & -  &&   180.0 \\
W2VV++ \cite{li2019w2vv++}           & 32.0 & 68.2 & 78.8 & - & -  &&    179.0   \\
HGR    \cite{chen2020fine}           & 35.1 & 73.5 & 83.5 & - & -  &&    192.1 \\
GPO    \cite{chen2021learning}   & 37.3 & 73.4 & 82.4 & - & -  && 193.1\\
Dual Encoding~\cite{dong2021dual}    & 36.8 & 73.6 & 83.7 & - & -  && 194.1  \\ [2pt]
RIVRL               & \uline{39.3} & \uline{76.0} & \textbf{85.1} & \textbf{2.0} & \uline{55.4}    &&  \uline{200.4} \\
RIVRL with BERT      & \textbf{40.0} & \textbf{76.4} & \uline{84.9} & \textbf{2.0} & \textbf{55.8}  && \textbf{201.3} \\

    \bottomrule
    \end{tabular}
 }
\end{table}

\subsubsection{Experiments on VATEX}
Table \ref{tab:vatex_perf} shows the performance on the VATEX dataset, where all
the models use the same I3D video feature, and do not pre-train on the HowTo100M dataset.
Among them, W2VV~\cite{dong2018predicting}, VSE++~\cite{fartash2017vse++} and W2VV++~\cite{li2019w2vv++} are all use mean pooling over the frame-level feature to encode videos, while our model explores multiple-branch video representation to encode videos, consistently achieving better performance. 
HGR~\cite{chen2020fine} and GPO~\cite{chen2021learning} utilize three independent branches to represent videos, but our proposed model using two dependent branches still performs better. The result to some extent shows the importance of using dependent multiple branches for video representation.
Additionally, Dual Encoding utilizes the same text encoder and hybrid space learning as ours, except for video representation. The superior performance of our model over Dual Encoding again demonstrates the benefit of using the reading-strategy inspired video representation for text-to-video retrieval.

\subsection{Analysis on Different Types of Videos} \label{ssec:speed}

\begin{figure}[tb!]
\centering
\subfigure[Grouped by video complexities]{
\includegraphics[width=\columnwidth]{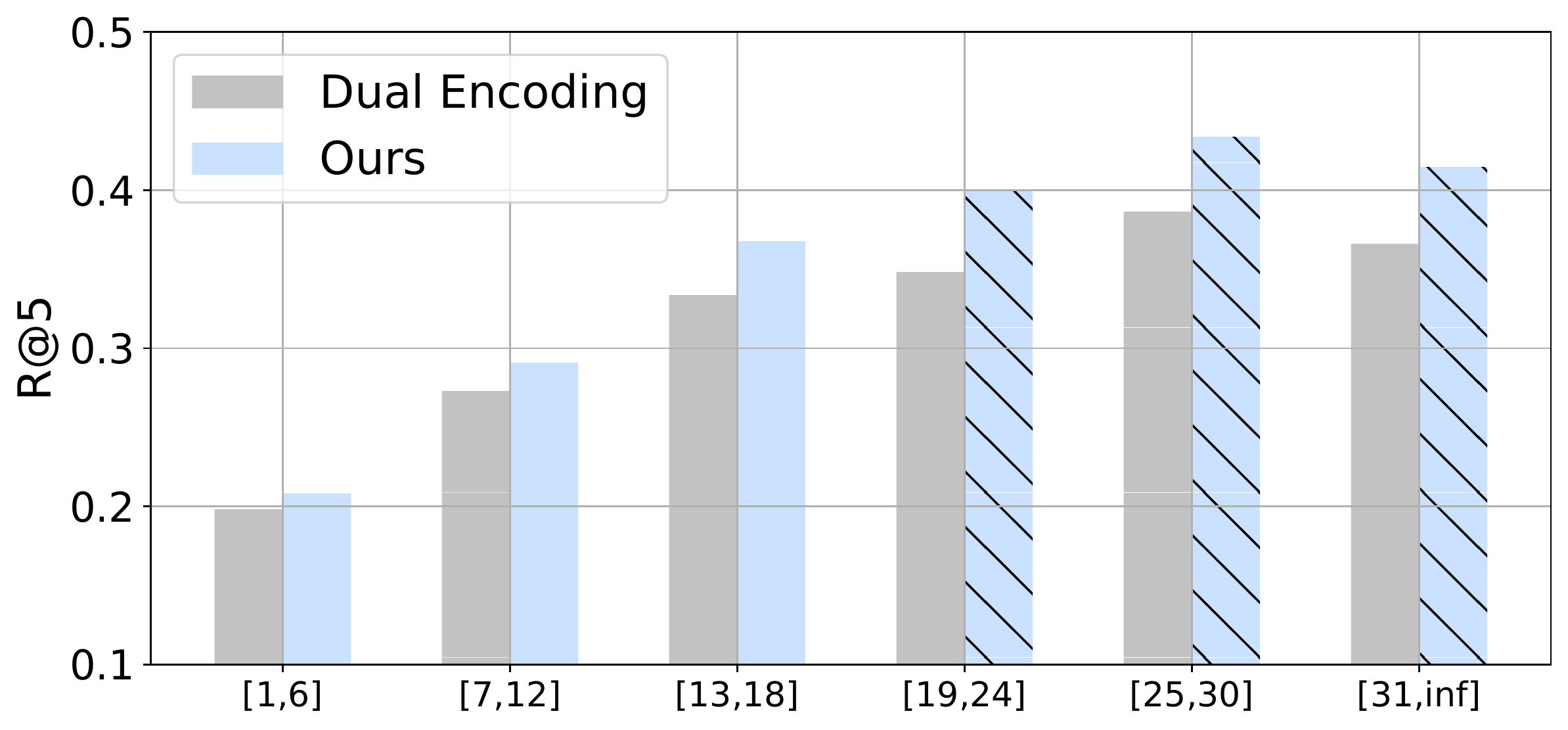}
}
\quad
\subfigure[Grouped by video categories]{
\includegraphics[width=\columnwidth]{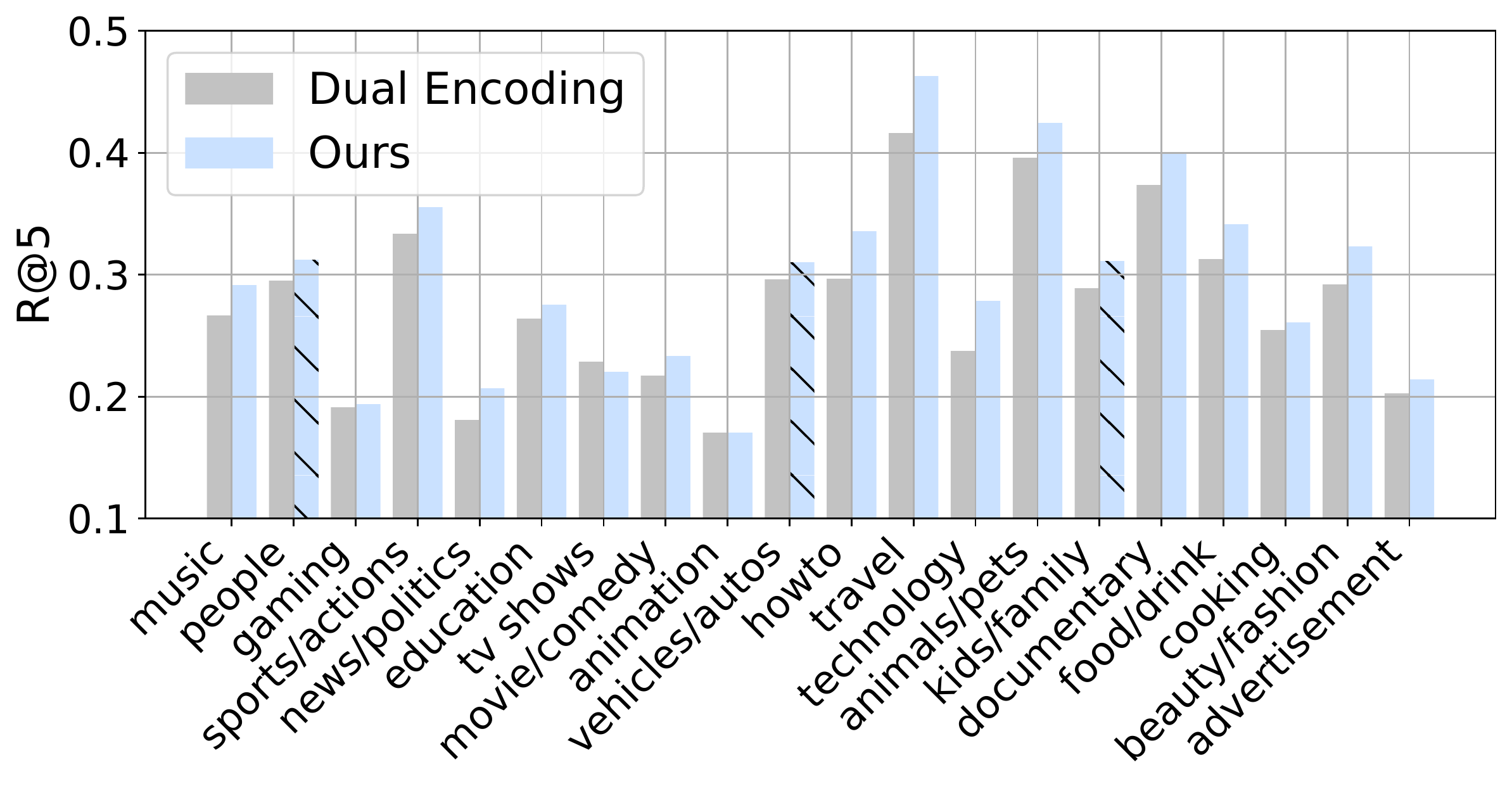}
}
\caption{\textbf{Performance comparison of our model and Dual Encoding~\cite{dong2021dual} on MV-Xu}. Queries have been grouped in terms of (a) complexities and (b) categories of the corresponding relevant videos. 
}\label{fig:showquery}
\end{figure}

\begin{figure*}[tb!]
\centering\includegraphics[width=2\columnwidth]{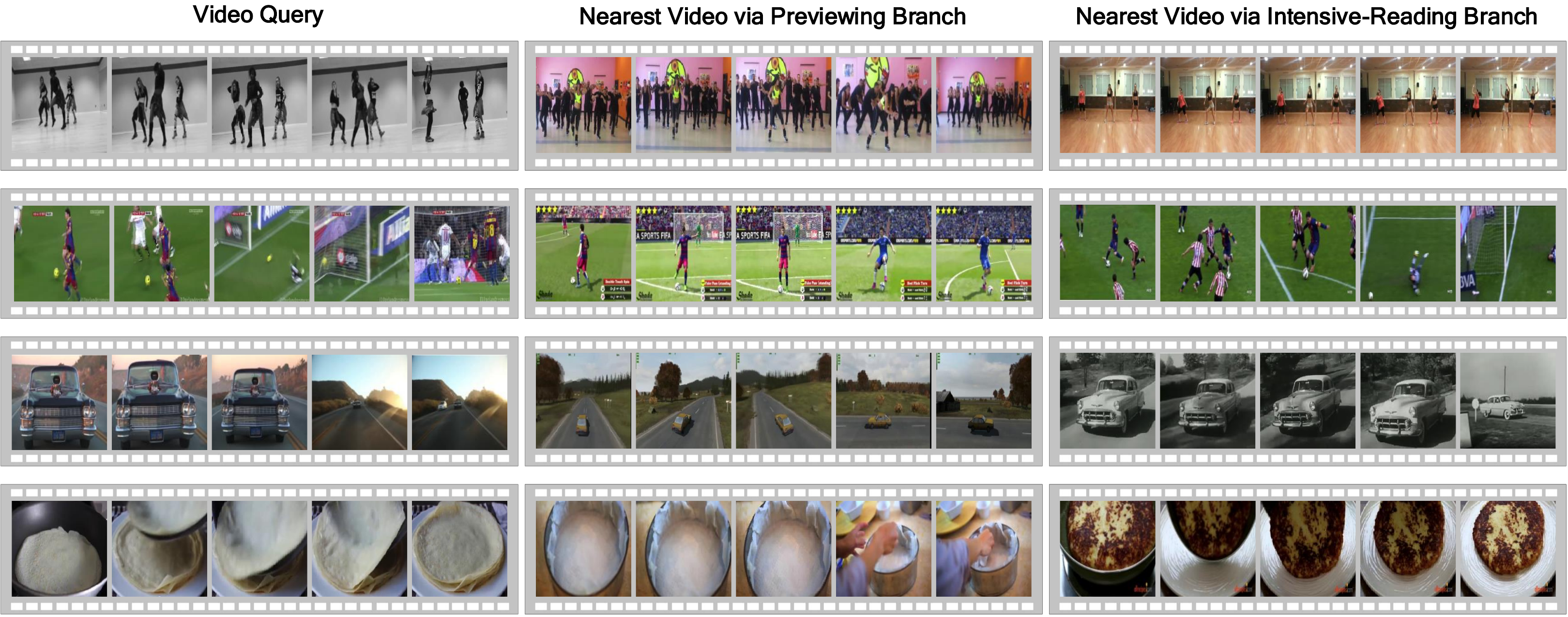}
\caption{\textbf{Examples of video-to-video retrieval on MSR-VTT}.  The previewing branch captures the coarse semantic relevance, while the intensive-reading branch has the ability to obtain more fine-grained semantic relevance. }\label{fig:video2video}
\vspace{-1mm}
\end{figure*}

\begin{figure*}[tb!]
\centering\includegraphics[width=2\columnwidth]{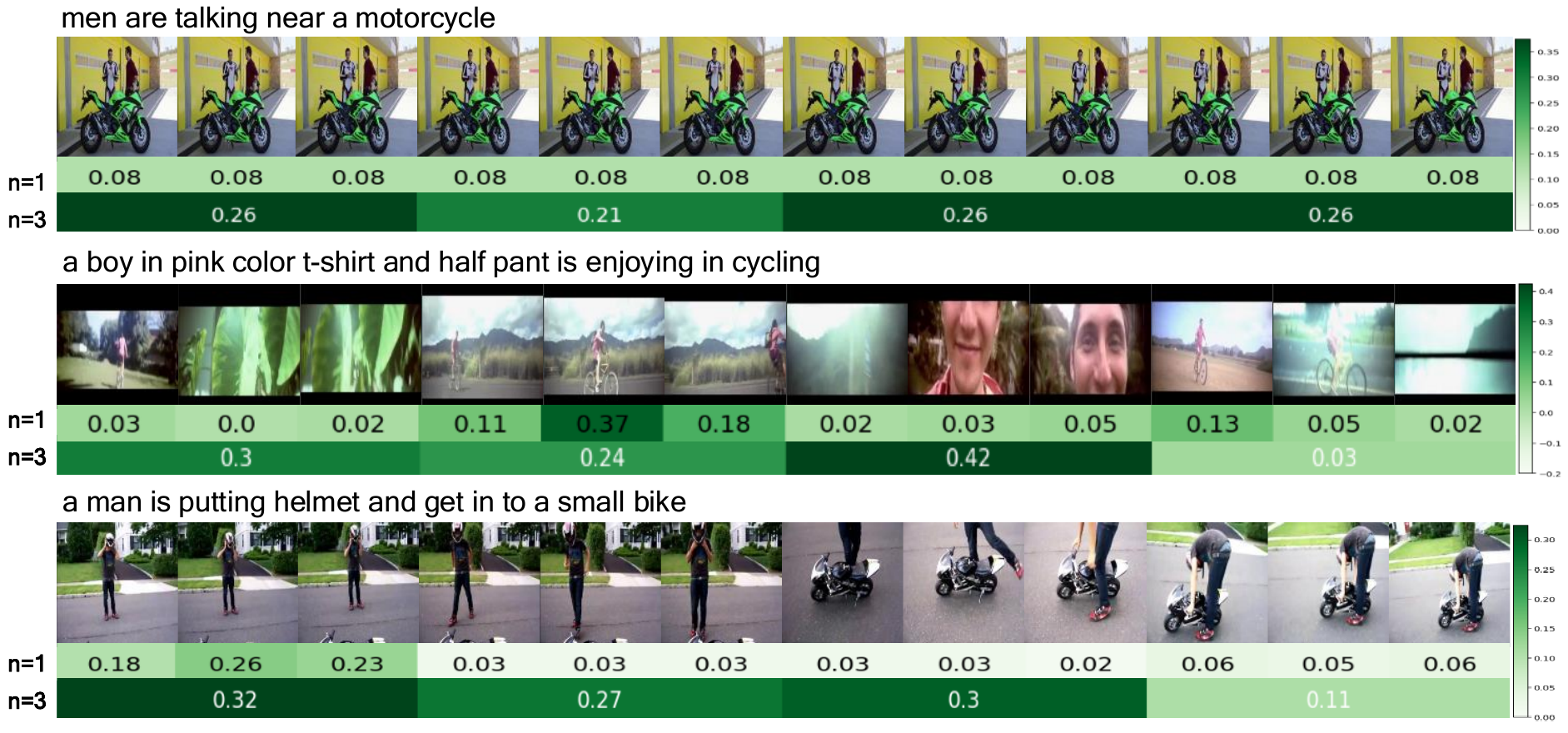}
\caption{\textbf{Visualization of previewing-aware attention}. The scores under videos are the attention weights obtained by our model, and darker color indicates a higher response of the corresponding frame (n=1) or segment (n=3).} 
\label{fig:attention_score}
\end{figure*}

In this section, we investigate how our proposed model performs on different types of videos, \ie video complexities and video categories. 
Notice that a video described by a longer sentence typically exhibits more complex visual content than another video described by a relatively shorter sentence. Hence, we use the length of corresponding textual descriptions as the metric of video complexity. A similar evaluation protocol is also adopted in~\cite{sigir2020tree}.
In order to make the comparison fairer, we compare our model with Dual Encoding that utilizes the same multi-level text encoding and hybrid space learning but uses different video representation methods.
Fig.~\ref{fig:showquery}(a) shows the performance comparison grouped in the variety of video complexities.
Our proposed model consistently outperforms Dual Encoding in all groups with a clear margin. The results again show the effectiveness of our proposed video representation.
Moreover, with the increasing of video complexities, the performance gain of our model over Dual Encoding becomes much more significant generally. The result also demonstrates that our reading-strategy inspired visual representation is better to handle complex videos.
Fig.~\ref{fig:showquery}(b) illustrates the performance comparison grouped in the variety of video categories.
Our proposed model beats Dual Encoding on almost all categories, which again shows the effectiveness of our proposed video representation.

\subsection{Qualitative Analysis}

\subsubsection{What has RIVRL learned}
In order to investigate what has RIVRL learned, we conduct video-to-video retrieval on MSR-VTT. Specifically, given a video, we search its nearest video using the features obtained by the previewing branch and the intensive-reading branch respectively, and the cosine distance is employed. Some examples are demonstrated in Fig.~\ref{fig:video2video}.
Considering the first example, given a query video containing \textit{three people are dancing}, the nearest video obtained by previewing branch is about \textit{a group of people are dancing}, while the one returned by the intensive-reading branch is also about \textit{three people are dancing}. 
For the second example, the one obtained by previewing branch shares the coarse semantic of \textit{football} with the query video, while the video obtained by intensive-reading branch shows a similar event of \textit{Messi dribbled and scored in a football match}.
The results show that the previewing branch captures the coarse semantic relevance, while the intensive-reading branch has the ability to obtain more fine-grained semantic relevance.

\subsubsection{Attention Visualization}
Fig. \ref{fig:attention_score} visualizes the learned attention of our proposed model. For each video, we select twelve frames for visualization and illustrate its frame-level attention (n=1) and segment-level attention (n=3).
Weights of each attention are L2-normalized for better visualization. Besides, one sentence of each video is also demonstrated for reference.
Considering the first video with a single scene where all the frames are almost similar, learned weights of frame-level attention and segment-level attention are also almost equal, respectively. 
For the video having multiple different scenes, our attention typically selects segments conveying the semantic of the video. Taking the second video as an example, the frame-level attention focuses more on the scene of \textit{a boy cycling}, while the segment-level attention focuses on the scene of \textit{a boy is enjoying}. The results to some extent demonstrate the reasonability of our learned attention.

\subsection{Ablation Studies} \label{ssec:ablation}

In this section, we evaluate the effectiveness of each component in our model. 
We first verify the viability of two-branch video representation, followed by the exploration of multi-granularity representation encoding, multi-head attention, and dependency  modeling.
Following previous works~\cite{luo2020univl}\cite{patrick2020support}\cite{bain2021frozen}, we conduct ablation studies on MV-Yu data split.

\begin{table} [tb!]
\renewcommand{\arraystretch}{1.2}
\caption{The effectiveness of two-branch video representation. Video representation obtained by both previewing branch and intensive-reading branch performs the best.}
\label{tab:two-branch}
\centering 
\scalebox{0.81}{
\begin{tabular}{ccl*{5}{r}c}
\toprule
\textbf{Previewing} & \textbf{Intensive-Reading} &  \textbf{R@1} & \textbf{R@5} & \textbf{R@10} & \textbf{Med r} & \textbf{mAP} & \textbf{SumR} \\
\cmidrule(l){1-2}\cmidrule(l){3-8}
\checkmark (biGRU)  &  -                &  20.4 & 47.6 & 59.9 & 7.0 & 33.2  & 127.9 \\
-  &  \checkmark                         &  22.1 & 48.5 & 60.9 & 6.0 & 34.8  & 131.5  \\
\checkmark (biGRU)  &  \checkmark       &  \textbf{23.3} & \textbf{52.2} & \textbf{63.8} & \textbf{5.0} & \textbf{36.7}  & \textbf{139.3}   \\
\cmidrule(l){1-2}\cmidrule(l){3-8}
\checkmark (FC)  &  -                &  18.9 & 42.7  &57.2  &7.5 &30.6  &118.8 \\
-  &  \checkmark                     &  22.1 & 48.5 & 60.9 & 6.0 & 34.8  & 131.5  \\
\checkmark (FC)  &  \checkmark       &  \textbf{22.3} & \textbf{51.3} & \textbf{61.5} & \textbf{5.0} & \textbf{35.4} & \textbf{135.1}  \\
\bottomrule
\end{tabular}
 }
\end{table}

\subsubsection{Single branch \textit{vs.} Multiple branches}
In order to verify the viability of two-branch video representation, we compare it with single-branch video representation, using a previewing branch or an intensive-reading branch only.
As shown in Table~\ref{tab:two-branch}, using biGRU as the previewing branch, our proposed two-branch video representation consistently outperforms two single-branch variants with a clear margin. Among the single-branch methods, the intensive-reading branch with a strong model structure unsurprisingly performs better. 
Additionally, we also try replacing the GRU as an FC layer as the previewing branch, and the two-branch method also performs the best. The results demonstrate the effectiveness of our proposed two-branch video representation.

\begin{table} [tb!]
\renewcommand{\arraystretch}{1.2}
\caption{The comparison of two dependent branches and two independent branches. 
Our model using two dependent branches for video representation consistently performs better. }
\label{tab:ablation-attention}
\centering 
\scalebox{0.99}{
\begin{tabular}{cc*{5}{r}c}
\toprule
\textbf{Dependent?}  &  \textbf{R@1} & \textbf{R@5} & \textbf{R@10} & \textbf{Med r} & \textbf{mAP} & \textbf{SumR} \\
\cmidrule(l){1-1} \cmidrule(l){2-7}
\Checkmark & \textbf{23.3}  & \textbf{52.2} & \textbf{63.8} & \textbf{5.0} & \textbf{36.7}  & \textbf{139.3} \\
\XSolidBrush & 23.0  & 50.9 & 62.6 & \textbf{5.0} & 36.0  & 136.5  \\
\bottomrule
\end{tabular}
 }
\end{table}

\subsubsection{Dependent branches \textit{vs.} Independent branches}
In this experiment, we compare our full model using two dependent branches to the counterpart using two independent branches. 
The former feeds the preview feature $p$ as the query Q of the multi-head attention in the intensive-reading branch, while the latter does not feed the previewing feature into the intensive-reading branch.
As shown in Table \ref{tab:ablation-attention}, our full model using two dependent branches gives better performance. 
The result demonstrates the benefit of using two dependent branches for video representation. We attribute the higher performance to that the preview feature provides a coarse overview of the video to the intensive-reading branch, thus the intensive-reading branch can extract essential information better according to the video overview.

\begin{table} [tb!]
\renewcommand{\arraystretch}{1.2}
\caption{The effectiveness of multi-granularity segment representation. Models with multi-granularity segment representation consistently outperform the single-granularity counterparts.}
\label{tab:multi-granularity}
\centering 
\scalebox{0.85}{
\begin{tabular}{ccl*{5}{r}c}
\toprule
\textbf{Granularity} & \textbf{Window size} &  \textbf{R@1} & \textbf{R@5} & \textbf{R@10} & \textbf{Med r} & \textbf{mAP} & \textbf{SumR} \\
\cmidrule(l){1-2} \cmidrule(l){3-8}
\multirow{3}{*}{Single}  & 1            & 22.0  & 50.4 & 63.5 & \textbf{5.0} & 35.6  & 135.9 \\
&  3             & 21.0  & 51.0 & 63.0 & \textbf{5.0} & 34.7  & 135.0   \\
&  5            & 22.3  & 49.7 & 61.5 & 6.0 & 35.5  & 133.5   \\
\cmidrule(l){1-2} \cmidrule(l){3-8}
\multirow{4}{*}{Multiple}  &  \{1,3\}          & \textbf{23.3}  & \textbf{52.2} & \textbf{63.8} & \textbf{5.0} & \textbf{36.7}  & \textbf{139.3}\\
&  \{1,5\}          & 22.8  & 51.7 & 63.2 & \textbf{5.0} & 36.2  & 137.7   \\
&  \{3,5\}          & 21.8  & 50.6 & 64.0 & \textbf{5.0} & 35.1  & 136.4   \\
&  \{1,3,5\}        & 22.0  & 51.2 & 63.7 & \textbf{5.0} & 35.7  & 136.9   \\
\bottomrule
\end{tabular}
 }
\end{table}

\subsubsection{The effectiveness of multi-granularity segment representation}
Table \ref{tab:multi-granularity} summarizes the performance of our proposed model with the different granularities of segment representation. The models with multiple granularities beat the single-granularity counterparts. The result shows the effectiveness of multi-granularity segment representation for video retrieval. 
Among multi-granularity models, one with sliding windows of sizes 1 and 3 turns out to be the most effective, but using more granularities has no further performance gain.
While its learning capacity increases as the granularity of the segment increases, the chance of over-fitting also increases.
Overall, our model using multi-granularity segment representation with sliding windows of sizes 1 and 3 strikes the best balance between model capacity and generalization ability, so we use it as the default setup of our model.

\begin{table} [tb!]
\renewcommand{\arraystretch}{1.2}
\caption{The effectiveness of multi-head attention. Our model using multi-head attention performs the best.
}
\label{tab:ablation-attention}
\centering 
\scalebox{0.93}{
\begin{tabular}{lc*{5}{r}c}
\toprule
\textbf{Attention}  &  \textbf{R@1} & \textbf{R@5} & \textbf{R@10} & \textbf{Med r} & \textbf{mAP} & \textbf{SumR} \\
\cmidrule(l){1-1} \cmidrule(l){2-7}
Mean pooling           & 20.5  & 48.3 & 61.1 & 6.0 & 33.5  & 129.9  \\
Simple attention       & 22.6  & 50.4 & 62.6 & 5.0 & 35.4  & 135.6  \\
Multi-head attention   & \textbf{23.3}  & \textbf{52.2} & \textbf{63.8} & \textbf{5.0} & \textbf{36.7}  & \textbf{139.3} \\
\bottomrule
\end{tabular}
 }
\end{table}

\subsubsection{The effectiveness of multi-head attention}
In this experiment, we try replacing the multi-head attention with mean pooling, or simple attention to verify its effectiveness in the previewing-aware attention. 
The mean pooling regards input segments equally, calculating the average of segment features.
The simple attention utilizes an FC layer to learn attention weights, where we first concatenate the preview feature with the input segment feature vectors, and an FC layer is employed to map each feature vector into an attention weight.
Two attentions are consistently better than the mean pooling, showing the benefit of attention in video representation.
Besides, we find that replacing the multi-head attention with the simple attention would result in relative performance degeneration, but not dramatically. It not only reflects the effectiveness of the multi-head attention, but also shows the robustness of our model structure.

\subsubsection{The influence of dependency modeling}

\begin{table} [tb!]
\renewcommand{\arraystretch}{0.9}
\caption{The comparison of different dependency  modelings. Our PaA outperforms the two ways used in SlowFast.}
\label{tab:ablation-fusion}
\centering 
\scalebox{0.95}{
\begin{tabular}{lc*{5}{r}c}
\toprule
\textbf{Lateral connection}  &  \textbf{R@1} & \textbf{R@5} & \textbf{R@10} & \textbf{Med r} & \textbf{mAP} & \textbf{SumR}  \\
\cmidrule(l){1-1} \cmidrule(l){2-7}
Concat     & 22.4  & 59.0 & 60.6 & 6.0 & 34.8  & 132.0  \\
Sum         & 21.7  & 48.3 & 61.4 & 6.0 & 34.7  & 133.3  \\
PaA         & \textbf{23.3}  & \textbf{52.2} & \textbf{63.8} & \textbf{5.0} & \textbf{36.7}  & \textbf{139.3} \\
\bottomrule
\end{tabular}
 }
\end{table}

Recall that we devise a previewing-aware attention to make the two branches dependent. In order to investigate its effectiveness, we compare it with two common ways that are used in the SlowFast network~\cite{feichtenhofer2019slowfast}, \ie summation and concatenation, for dependency modeling.
Specifically, instead of using the previewing-aware attention, we first fuse the representation of previewing branch $p$ and segment representation $C'$ by element-wise summation or concatenation, and a self attention depicted in Fig. \ref{fig:attention_pic}(a) followed with an average pooling layer is further employed.
Table \ref{tab:ablation-fusion} shows the result on the MSR-VTT dataset. Our proposed previewing-aware attention consistently outperforms the two ways used in SlowFast, demonstrating its effectiveness as the dependency modeling for text-to-video retrieval.

\subsection{Video-to-Video Retrieval} \label{ssec:v2v_retrieval}
\begin{table} [tb!]
\renewcommand{\arraystretch}{1.2}
\caption{The quantitative comparison of video-to-video retrieval. The performance are reported in percentage (\%).}
\label{tab:v2v-retrieval}
\centering 
\scalebox{1}{
\begin{tabular}{lc*{5}{r}c}
\toprule
\textbf{Representation}  & \textbf{nDCG} & \textbf{mAP} \\
\cmidrule(l){1-1} \cmidrule(l){2-3}
Baseline & 59.6 & 9.1\\
Previewing branch & 65.9 & 10.8\\
Intensive-reading branch & 66.3 & 11.0\\
RIVRL & \textbf{70.0} & \textbf{16.8 }\\
\bottomrule
\end{tabular}
 }
\end{table}

To further verify the effectiveness of our learned video representation, we conduct a quantitative experiment of video-to-video retrieval on MSR-VTT. 
Considering the MSR-VTT dataset can not be directly used for the video-to-video retrieval experiment, we generate video-to-video annotations by measuring sentence-to-sentence similarities between associated sentences of the corresponding videos. 
In particular, given two videos associated with 20 sentences respectively, we are able to obtain $20\times20=400$ sentence pairs between two videos and measure their sentence-to-sentence similarities by synset-aware semantic similarity proposed in~\cite{wray2021semantic}.
The average similarity of 400 sentence pairs is regarded as the corresponding video-to-video similarity. We regard two videos as relevant if their video-to-video similarity is larger than 0.2, otherwise irrelevant.
Based on the above annotation process, we conduct experiments on the test set of 2,990 videos, where videos that have at least one relevant video are deemed as the query videos, resulting in 308 queries.
The quantitative results are shown in \ref{tab:v2v-retrieval}, where two common retrieval metrics \ie nDCG and mAP, are used for performance evaluation. Besides, all methods utilize cosine similarity between video features for similarity computation.
The lower three methods via representation learning are significantly superior to the baseline which directly utilize the mean-pooled ResNeXt-ResNet feature without learning on MSR-VTT. The results show the necessity of representation learning for video-to-video retrieval. Additionally, among the three methods via representation learning, our RIVRL gives the best performance, demonstrating the effectiveness of two dependent branches for video representation.

\subsection{Analysis on Model Complexity} \label{ssec:complexity}

\begin{table}[tbp!]
	\centering
	\caption{Complexity comparison in terms of model size and computation overhead at the inference stage. Lower scores indicate better.}
	\label{tab:model-complexity}
	\begin{tabular}{lrr}
		\toprule  
	\multirow{2}{*}{\textbf{Model}}   &\multicolumn{2}{c}{\textbf{Model Complexity}} \\ 
		\cmidrule(r){2-3}
		  &Parameters(M) &FLOPs(G)\\
		  \cmidrule{1-3}
		  CE \cite{liu2019use}                  & 183.45 & 20.16\\[2pt]
		  Dual Encoding \cite{dong2021dual}     & 32.40 & 1.36\\[2pt]
		  MMT \cite{gabeur2020multi}             & 133.40 & 12.64\\ [2pt]
		  RIVRL                      & 113.37 & 2.50\\
		\bottomrule  
	\end{tabular}
\end{table}

In this section, we compare our model with recent state-of-the-art models in terms of model size and computation overhead at the inference stage. 
We adopt two non-Transformer models, CE \cite{liu2019use} and Dual Encoding \cite{dong2021dual}, and a Transformer-based model MMT~\cite{gabeur2020multi}, considering their state-of-the-art performance and open-source code.
Specifically, for each model, we measure model size and the number of FLOPs it takes to encode a given video-text pair via an open-source toolbox\footnote{https://github.com/sovrasov/flops-counter.pytorch}. 
The computational cost of video feature extraction is excluded as that step is typically performed once in an offline manner.
As shown in Table \ref{tab:model-complexity}, our proposed model is worse than Dual Encoding, but much better than CE that uses seven different video features and MMT based on heavy Transformers.
Note that the most state-of-art method T2VLAD~\cite{wang2021t2vlad} is not included, as this method remains closed-source, making a precise estimation of its model complexity impossible. However, as T2VLAD employs an extra local branch on basis of CE, it is much more complicated than CE and accordingly also more complicated than ours.

\section{Summary and Conclusions} \label{sec:conc}

Inspired by the reading strategy of humans, we have presented a new model by exploiting two-branch video representation for text-to-video retrieval.
In our model, a previewing branch and an intensive-reading branch are jointly learned to represent videos.
Extensive experiments on three datasets, \ie MSR-VTT, TGIF, VATEX, demonstrate the effectiveness of our proposed model for text-to-video retrieval. Besides, compared to heavy Transformer based models, our model shows a good balance between retrieval performance and computational complexity.
We also find that our two-branch video representation architecture is insensitive to attention implementation and enjoys good robustness.
Benefited from the reading-strategy inspired visual representation learning, our proposed model also shows its potential for handling complex videos.

\appendices

{
\section{Video-Text Similarity in Hybrid Space}
As a hybrid space consists of a latent space and a concept space, the video-text similarity in the hybrid space is computed as the weighted sum of their similarity in the latent space and the concept space.
Given a video represented as $v$ and a sentence represented as $s$, their video-text similarity in the hybrid space is calculated as:
\begin{equation}
sim_{hyb}(v,s) = \alpha \cdot sim_{lat}(v,s) + (1-\alpha) \cdot sim_{con}(v,s),
\end{equation}
where $\alpha$ is a hyper-parameter to balance the importance of two spaces. Following \cite{dong2021dual}, we set $\alpha$ to be 0.6. Besides, we use the cosine similarity to measure the video-text similarity $sim_{lat}(v,s)$ in the latent space, and the generalized Jaccard similarity to compute the video-text similarity $sim_{con}(v,s)$ in the concept space. Formally, the video-text similarity $sim_{lat}(v,s)$ in the latent space is calculated as:
\begin{equation}
    sim_{lat}(v,s) = \frac{f(v) \cdot f(s)}{\left\|f(v)\right\| \left\|f(s)\right\|},
\end{equation}
where $f(v)$ and $f(s)$ indicate the video feature vector and sentence feature vector in the latent space, respectively.
Additionally, the video-text similarity $sim_{con}(v,s)$ in the concept space is calculated as:
\begin{equation}
sim_{con}(v, s) = \frac{\sum_{i=1}^{K} \min(g(v)_i,g(s)_i)}{\sum_{i=1}^{K} \max(g(v)_i,g(s)_i)},
\end{equation}
where $K$ denotes the dimensionaity of the concept space. $g(v)$ indicates the video feature vector in the concept space, which is actually the probability vector of $K$ concepts being relevant with respect to the video $v$. $g(v)_i$ is the $i$-th element in $g(v)$. Similarly, $g(s)_i$ is the $i$-th element in the sentence feature vector $g(s)$ of the sentence $s$ in the concept space.

Based on the above computation of video-text similarity in hybrid space, the video-text similarity $sim_{p}(v,s)$ in the previewing hybrid space is calculated on the video representation $p$ obtained by the previewing branch and the sentence representation $s$, namely
\begin{equation}
sim_p(v,s) = sim_{hyb}(p,s).
\end{equation}
Similarly, the video-text similarity $sim_{g}(v,s)$ in the intensive hybrid space is calculated on the video representation $g$ obtained by the intensive-reading branch and the sentence representation $s$, namely
\begin{equation}
sim_g(v,s) = sim_{hyb}(g,s).
\end{equation}

}

\ifCLASSOPTIONcaptionsoff
  \newpage
\fi



\bibliographystyle{IEEEtran}
\bibliography{IEEEabrv,referemce}

\begin{thebibliography}{10}
\providecommand{\url}[1]{#1}
\csname url@samestyle\endcsname
\providecommand{\newblock}{\relax}
\providecommand{\bibinfo}[2]{#2}
\providecommand{\BIBentrySTDinterwordspacing}{\spaceskip=0pt\relax}
\providecommand{\BIBentryALTinterwordstretchfactor}{4}
\providecommand{\BIBentryALTinterwordspacing}{\spaceskip=\fontdimen2\font plus
\BIBentryALTinterwordstretchfactor\fontdimen3\font minus
  \fontdimen4\font\relax}
\providecommand{\BIBforeignlanguage}[2]{{%
\expandafter\ifx\csname l@#1\endcsname\relax
\typeout{** WARNING: IEEEtran.bst: No hyphenation pattern has been}%
\typeout{** loaded for the language `#1'. Using the pattern for}%
\typeout{** the default language instead.}%
\else
\language=\csname l@#1\endcsname
\fi
#2}}
\providecommand{\BIBdecl}{\relax}
\BIBdecl

\bibitem{huang2021holographic}
Y.~Huang, X.~Yang, J.~Gao, and C.~Xu, ``Holographic feature learning of
  egocentric-exocentric videos for multi-domain action recognition,''
  \emph{IEEE Transactions on Multimedia}, 2021.

\bibitem{wu2021spatiotemporal}
H.~Wu, X.~Ma, and Y.~Li, ``Spatiotemporal multimodal learning with 3d cnns for
  video action recognition,'' \emph{IEEE Transactions on Circuits and Systems
  for Video Technology}, 2021.

\bibitem{tan2021selective}
Y.~Tan, Y.~Hao, X.~He, Y.~Wei, and X.~Yang, ``Selective dependency aggregation
  for action classification,'' in \emph{Proceedings of the 29th ACM
  International Conference on Multimedia}, 2021, pp. 592--601.

\bibitem{li2021interventional}
Y.~Li, X.~Yang, X.~Shang, and T.-S. Chua, ``Interventional video relation
  detection,'' in \emph{Proceedings of the 29th ACM International Conference on
  Multimedia}, 2021, pp. 4091--4099.

\bibitem{xu2018dual}
N.~Xu, A.-A. Liu, Y.~Wong, Y.~Zhang, W.~Nie, Y.~Su, and M.~Kankanhalli,
  ``Dual-stream recurrent neural network for video captioning,'' \emph{IEEE
  Transactions on Circuits and Systems for Video Technology}, vol.~29, no.~8,
  pp. 2482--2493, 2018.

\bibitem{deng2021syntax}
J.~Deng, L.~Li, B.~Zhang, S.~Wang, Z.~Zha, and Q.~Huang, ``Syntax-guided
  hierarchical attention network for video captioning,'' \emph{IEEE
  Transactions on Circuits and Systems for Video Technology}, 2021.

\bibitem{shang2019annotating}
X.~Shang, D.~Di, J.~Xiao, Y.~Cao, X.~Yang, and T.-S. Chua, ``Annotating objects
  and relations in user-generated videos,'' in \emph{Proceedings of the 2019 on
  International Conference on Multimedia Retrieval}, 2019, pp. 279--287.

\bibitem{dong2016early}
J.~Dong, X.~Li, W.~Lan, Y.~Huo, and C.~G. Snoek, ``Early embedding and late
  reranking for video captioning,'' in \emph{Proceedings of the 24th ACM
  international conference on Multimedia}, 2016, pp. 1082--1086.

\bibitem{araujo2017large}
A.~Araujo and B.~Girod, ``Large-scale video retrieval using image queries,''
  \emph{IEEE Transactions on Circuits and Systems for Video Technology},
  vol.~28, no.~6, pp. 1406--1420, 2017.

\bibitem{gao2021learning}
J.~Gao and C.~Xu, ``Learning video moment retrieval without a single annotated
  video,'' \emph{IEEE Transactions on Circuits and Systems for Video
  Technology}, 2021.

\bibitem{yang2021deconfounded}
X.~Yang, F.~Feng, W.~Ji, M.~Wang, and T.-S. Chua, ``Deconfounded video moment
  retrieval with causal intervention,'' in \emph{Proceedings of the 44th
  International ACM SIGIR Conference on Research and Development in Information
  Retrieval}, 2021, pp. 1--10.

\bibitem{yang2022video}
X.~Yang, S.~Wang, J.~Dong, J.~Dong, M.~Wang, and T.-S. Chua, ``Video moment
  retrieval with cross-modal neural architecture search,'' \emph{IEEE
  Transactions on Image Processing}, vol.~31, pp. 1204--1216, 2022.

\bibitem{cikm13-zsvr}
J.~Dalton, J.~Allan, and P.~Mirajkar, ``Zero-shot video retrieval using content
  and concepts,'' in \emph{Proceedings of the 22nd ACM international conference
  on Information \& Knowledge Management}, 2013, pp. 1857--1860.

\bibitem{aaai15-zsed}
X.~Chang, Y.~Yang, A.~Hauptmann, E.~P. Xing, and Y.-L. Yu, ``Semantic concept
  discovery for large-scale zero-shot event detection,'' in \emph{Twenty-fourth
  international joint conference on artificial intelligence}, 2015, pp.
  2234--2240.

\bibitem{icmr2017-certh-avs}
F.~Markatopoulou, D.~Galanopoulos, V.~Mezaris, and I.~Patras, ``Query and
  keyframe representations for ad-hoc video search,'' in \emph{Proceedings of
  the ACM on International Conference on Multimedia Retrieval}, 2017, pp.
  407--411.

\bibitem{pami2017-videostory}
A.~Habibian, T.~Mensink, and C.~G. Snoek, ``Video2vec embeddings recognize
  events when examples are scarce,'' \emph{IEEE Transactions on Pattern
  Analysis and Machine Intelligence}, vol.~39, no.~10, pp. 2089--2103, 2016.

\bibitem{wang2020learning}
W.~Wang, J.~Gao, X.~Yang, and C.~Xu, ``Learning coarse-to-fine graph neural
  networks for video-text retrieval,'' \emph{IEEE Transactions on Multimedia},
  2020.

\bibitem{song2021spatial}
X.~Song, J.~Chen, Z.~Wu, and Y.-G. Jiang, ``Spatial-temporal graphs for
  cross-modal text2video retrieval,'' \emph{IEEE Transactions on Multimedia},
  2021.

\bibitem{yang2020weakly}
X.~Yang, X.~Liu, M.~Jian, X.~Gao, and M.~Wang, ``Weakly-supervised video object
  grounding by exploring spatio-temporal contexts,'' in \emph{Proceedings of
  the 28th ACM international conference on multimedia}, 2020, pp. 1939--1947.

\bibitem{wang2021t2vlad}
X.~Wang, L.~Zhu, and Y.~Yang, ``T2vlad: global-local sequence alignment for
  text-video retrieval,'' in \emph{Proceedings of the IEEE/CVF Conference on
  Computer Vision and Pattern Recognition}, 2021, pp. 5079--5088.

\bibitem{he2021improving}
F.~He, Q.~Wang, Z.~Feng, W.~Jiang, Y.~L{\"u}, Y.~Zhu, and X.~Tan, ``Improving
  video retrieval by adaptive margin,'' in \emph{Proceedings of the 44th
  International ACM SIGIR Conference on Research and Development in Information
  Retrieval}, 2021, pp. 1359--1368.

\bibitem{xiao2020visual}
J.~Xiao, X.~Shang, X.~Yang, S.~Tang, and T.-S. Chua, ``Visual relation
  grounding in videos,'' in \emph{European Conference on Computer
  Vision}.\hskip 1em plus 0.5em minus 0.4em\relax Springer, 2020, pp. 447--464.

\bibitem{dong2019dual}
J.~Dong, X.~Li, C.~Xu, S.~Ji, Y.~He, G.~Yang, and X.~Wang, ``Dual encoding for
  zero-example video retrieval,'' in \emph{Proceedings of the IEEE/CVF
  Conference on Computer Vision and Pattern Recognition}, 2019, pp. 9346--9355.

\bibitem{eccv2016ws-otani}
M.~Otani, Y.~Nakashima, E.~Rahtu, J.~Heikkil{\"a}, and N.~Yokoya, ``Learning
  joint representations of videos and sentences with web image search,'' in
  \emph{Proceedings of the European Conference on Computer Vision}, 2016, pp.
  651--667.

\bibitem{mithun2018learning}
N.~C. Mithun, J.~Li, F.~Metze, and A.~K. Roy-Chowdhury, ``Learning joint
  embedding with multimodal cues for cross-modal video-text retrieval,'' in
  \emph{Proceedings of the 2018 ACM on International Conference on Multimedia
  Retrieval}, 2018, pp. 19--27.

\bibitem{wray2019fine}
M.~Wray, D.~Larlus, G.~Csurka, and D.~Damen, ``Fine-grained action retrieval
  through multiple parts-of-speech embeddings,'' in \emph{Proceedings of the
  IEEE/CVF International Conference on Computer Vision}, 2019, pp. 450--459.

\bibitem{miech2019howto100m}
A.~Miech, D.~Zhukov, J.-B. Alayrac, M.~Tapaswi, I.~Laptev, and J.~Sivic,
  ``Howto100m: Learning a text-video embedding by watching hundred million
  narrated video clips,'' in \emph{Proceedings of the IEEE/CVF International
  Conference on Computer Vision}, 2019, pp. 2630--2640.

\bibitem{torabi2016learning}
A.~Torabi, N.~Tandon, and L.~Sigal, ``Learning language-visual embedding for
  movie understanding with natural-language,'' \emph{arXiv preprint
  arXiv:1609.08124}, 2016.

\bibitem{dongdl}
J.~Dong, S.~Huang, D.~Xu, and D.~Tao, ``Dl-61-86 at trecvid 2017: Video-to-text
  description,'' in \emph{TRECVID Workshop}, 2017.

\bibitem{sigir2020tree}
X.~Yang, J.~Dong, Y.~Cao, X.~Wang, M.~Wang, and T.-S. Chua, ``Tree-augmented
  cross-modal encoding for complex-query video retrieval,'' in
  \emph{Proceedings of the 43rd International ACM SIGIR Conference on Research
  and Development in Information Retrieval}, 2020, pp. 1339--1348.

\bibitem{vaswani2017attention}
A.~Vaswani, N.~Shazeer, N.~Parmar, J.~Uszkoreit, L.~Jones, A.~N. Gomez,
  {\L}.~Kaiser, and I.~Polosukhin, ``Attention is all you need,'' in
  \emph{Advances on Neural Information Processing Systems}, 2017, pp.
  5998--6008.

\bibitem{gabeur2020multi}
V.~Gabeur, C.~Sun, K.~Alahari, and C.~Schmid, ``Multi-modal transformer for
  video retrieval,'' in \emph{Proceedings of the European Conference on
  Computer Vision}, vol.~5, 2020, pp. 214--229.

\bibitem{zhu2020actbert}
L.~Zhu and Y.~Yang, ``Actbert: Learning global-local video-text
  representations,'' in \emph{Proceedings of the IEEE/CVF Conference on
  Computer Vision and Pattern Recognition}, 2020, pp. 8746--8755.

\bibitem{miech2018learning}
A.~Miech, I.~Laptev, and J.~Sivic, ``Learning a text-video embedding from
  incomplete and heterogeneous data,'' \emph{arXiv preprint arXiv:1804.02516},
  2018.

\bibitem{li2020sea}
X.~Li, F.~Zhou, C.~Xu, J.~Ji, and G.~Yang, ``Sea: Sentence encoder assembly for
  video retrieval by textual queries,'' \emph{IEEE Transactions on Multimedia},
  2020.

\bibitem{chen2020fine}
S.~Chen, Y.~Zhao, Q.~Jin, and Q.~Wu, ``Fine-grained video-text retrieval with
  hierarchical graph reasoning,'' in \emph{Proceedings of the IEEE/CVF
  Conference on Computer Vision and Pattern Recognition}, 2020, pp.
  10\,638--10\,647.

\bibitem{song2019polysemous}
Y.~Song and M.~Soleymani, ``Polysemous visual-semantic embedding for
  cross-modal retrieval,'' in \emph{Proceedings of the IEEE/CVF Conference on
  Computer Vision and Pattern Recognition}, 2019, pp. 1979--1988.

\bibitem{cho2014learning}
K.~Cho, B.~Van~Merri{\"e}nboer, C.~Gulcehre, D.~Bahdanau, F.~Bougares,
  H.~Schwenk, and Y.~Bengio, ``Learning phrase representations using rnn
  encoder-decoder for statistical machine translation,'' \emph{arXiv preprint
  arXiv:1406.1078}, 2014.

\bibitem{ging2020coot}
S.~Ging, M.~Zolfaghari, H.~Pirsiavash, and T.~Brox, ``Coot: Cooperative
  hierarchical transformer for video-text representation learning,'' 2020.

\bibitem{dong2021dual}
J.~Dong, X.~Li, C.~Xu, X.~Yang, G.~Yang, X.~Wang, and M.~Wang, ``Dual encoding
  for video retrieval by text,'' \emph{IEEE Transactions on Pattern Analysis
  and Machine Intelligence}, 2021.

\bibitem{luo2021coco}
J.~Luo, Y.~Li, Y.~Pan, T.~Yao, H.~Chao, and T.~Mei, ``Coco-bert: Improving
  video-language pre-training with contrastive cross-modal matching and
  denoising,'' in \emph{Proceedings of the 29th ACM International Conference on
  Multimedia}, 2021, pp. 5600--5608.

\bibitem{liu2019use}
Y.~Liu, S.~Albanie, A.~Nagrani, and A.~Zisserman, ``Use what you have: Video
  retrieval using representations from collaborative experts,'' \emph{arXiv
  preprint arXiv:1907.13487}, 2019.

\bibitem{xu2016msr}
J.~Xu, T.~Mei, T.~Yao, and Y.~Rui, ``{MSR-VTT}: {A} large video description
  dataset for bridging video and language,'' in \emph{Proceedings of the
  IEEE/CVF conference on Computer Vision and Pattern Recognition}, 2016, pp.
  5288--5296.

\bibitem{tgif}
Y.~Li, Y.~Song, L.~Cao, J.~Tetreault, L.~Goldberg, A.~Jaimes, and J.~Luo,
  ``Tgif: A new dataset and benchmark on animated gif description,'' in
  \emph{Proceedings of the IEEE/CVF conference on Computer Vision and Pattern
  Recognition}, 2016, pp. 4641--4650.

\bibitem{wang2019vatex}
X.~Wang, J.~Wu, J.~Chen, L.~Li, Y.-F. Wang, and W.~Y. Wang, ``Vatex: A
  large-scale, high-quality multilingual dataset for video-and-language
  research,'' in \emph{Proceedings of the IEEE/CVF International Conference on
  Computer Vision}, 2019, pp. 4581--4591.

\bibitem{feng2020exploiting}
Z.~Feng, Z.~Zeng, C.~Guo, and Z.~Li, ``Exploiting visual semantic reasoning for
  video-text retrieval,'' \emph{arXiv preprint arXiv:2006.08889}, 2020.

\bibitem{arandjelovic2016netvlad}
R.~Arandjelovic, P.~Gronat, A.~Torii, T.~Pajdla, and J.~Sivic, ``Netvlad: Cnn
  architecture for weakly supervised place recognition,'' in \emph{Proceedings
  of the IEEE/CVF conference on Computer Vision and Pattern Recognition}, 2016,
  pp. 5297--5307.

\bibitem{dong2018predicting}
J.~Dong, X.~Li, and C.~G. Snoek, ``Predicting visual features from text for
  image and video caption retrieval,'' \emph{IEEE Transactions on Multimedia},
  vol.~20, no.~12, pp. 3377--3388, 2018.

\bibitem{shao2018find}
D.~Shao, Y.~Xiong, Y.~Zhao, Q.~Huang, Y.~Qiao, and D.~Lin, ``Find and focus:
  Retrieve and localize video events with natural language queries,'' in
  \emph{Proceedings of the European Conference on Computer Vision}, 2018, pp.
  200--216.

\bibitem{li2020hero}
L.~Li, Y.-C. Chen, Y.~Cheng, Z.~Gan, L.~Yu, and J.~Liu, ``Hero: Hierarchical
  encoder for video+ language omni-representation pre-training,'' \emph{arXiv
  preprint arXiv:2005.00200}, 2020.

\bibitem{chen2021learning}
J.~Chen, H.~Hu, H.~Wu, Y.~Jiang, and C.~Wang, ``Learning the best pooling
  strategy for visual semantic embedding,'' in \emph{Proceedings of the
  IEEE/CVF Conference on Computer Vision and Pattern Recognition}, 2021, pp.
  15\,789--15\,798.

\bibitem{carreira2017quo}
J.~Carreira and A.~Zisserman, ``Quo vadis, action recognition? a new model and
  the kinetics dataset,'' in \emph{Proceedings of the IEEE/CVF conference on
  Computer Vision and Pattern Recognition}, 2017, pp. 6299--6308.

\bibitem{liu2020sibnet}
S.~Liu, Z.~Ren, and J.~Yuan, ``Sibnet: Sibling convolutional encoder for video
  captioning,'' \emph{IEEE transactions on pattern analysis and machine
  intelligence}, 2020.

\bibitem{feichtenhofer2019slowfast}
C.~Feichtenhofer, H.~Fan, J.~Malik, and K.~He, ``Slowfast networks for video
  recognition,'' in \emph{Proceedings of the IEEE/CVF international conference
  on computer vision}, 2019, pp. 6202--6211.

\bibitem{dong2021multi}
J.~Dong, Z.~Long, X.~Mao, C.~Lin, Y.~He, and S.~Ji, ``Multi-level alignment
  network for domain adaptive cross-modal retrieval,'' \emph{Neurocomputing},
  vol. 440, pp. 207--219, 2021.

\bibitem{yu2018joint}
Y.~Yu, J.~Kim, and G.~Kim, ``A joint sequence fusion model for video question
  answering and retrieval,'' in \emph{Proceedings of the European Conference on
  Computer Vision}, 2018, pp. 471--487.

\bibitem{lei2021less}
J.~Lei, L.~Li, L.~Zhou, Z.~Gan, T.~L. Berg, M.~Bansal, and J.~Liu, ``Less is
  more: Clipbert for video-and-language learning via sparse sampling,'' in
  \emph{Proceedings of the IEEE/CVF Conference on Computer Vision and Pattern
  Recognition}, 2021, pp. 7331--7341.

\bibitem{radford2021learning}
A.~Radford, J.~W. Kim, C.~Hallacy, A.~Ramesh, G.~Goh, S.~Agarwal, G.~Sastry,
  A.~Askell, P.~Mishkin, J.~Clark \emph{et~al.}, ``Learning transferable visual
  models from natural language supervision,'' \emph{arXiv preprint
  arXiv:2103.00020}, 2021.

\bibitem{yu2019deep}
Z.~Yu, J.~Yu, Y.~Cui, D.~Tao, and Q.~Tian, ``Deep modular co-attention networks
  for visual question answering,'' in \emph{Proceedings of the IEEE/CVF
  Conference on Computer Vision and Pattern Recognition}, 2019, pp. 6281--6290.

\bibitem{liu2021multimodal}
Y.~Liu, J.~Zhang, L.~Fang, Q.~Jiang, and B.~Zhou, ``Multimodal motion
  prediction with stacked transformers,'' in \emph{Proceedings of the IEEE/CVF
  Conference on Computer Vision and Pattern Recognition}, 2021, pp. 7577--7586.

\bibitem{hu2020iterative}
R.~Hu, A.~Singh, T.~Darrell, and M.~Rohrbach, ``Iterative answer prediction
  with pointer-augmented multimodal transformers for textvqa,'' in
  \emph{Proceedings of the IEEE/CVF Conference on Computer Vision and Pattern
  Recognition}, 2020, pp. 9992--10\,002.

\bibitem{yu2019multimodal}
J.~Yu, J.~Li, Z.~Yu, and Q.~Huang, ``Multimodal transformer with multi-view
  visual representation for image captioning,'' \emph{IEEE transactions on
  circuits and systems for video technology}, vol.~30, no.~12, pp. 4467--4480,
  2019.

\bibitem{wu2019liteeval}
Z.~Wu, C.~Xiong, Y.-G. Jiang, and L.~S. Davis, ``Liteeval: A coarse-to-fine
  framework for resource efficient video recognition,'' \emph{arXiv preprint
  arXiv:1912.01601}, 2019.

\bibitem{wu2019adaframe}
Z.~Wu, C.~Xiong, C.-Y. Ma, R.~Socher, and L.~S. Davis, ``Adaframe: Adaptive
  frame selection for fast video recognition,'' in \emph{Proceedings of the
  IEEE/CVF Conference on Computer Vision and Pattern Recognition}, 2019, pp.
  1278--1287.

\bibitem{chen2020image}
Y.~Chen, S.~Gong, and L.~Bazzani, ``Image search with text feedback by
  visiolinguistic attention learning,'' in \emph{Proceedings of the IEEE/CVF
  Conference on Computer Vision and Pattern Recognition}, 2020, pp. 3001--3011.

\bibitem{zhong2020self}
H.~Zhong, J.~Chen, C.~Shen, H.~Zhang, J.~Huang, and X.-S. Hua, ``Self-adaptive
  neural module transformer for visual question answering,'' \emph{IEEE
  Transactions on Multimedia}, vol.~23, pp. 1264--1273, 2020.

\bibitem{devlin2018bert}
J.~Devlin, M.-W. Chang, K.~Lee, and K.~Toutanova, ``Bert: Pre-training of deep
  bidirectional transformers for language understanding,'' \emph{arXiv preprint
  arXiv:1810.04805}, 2018.

\bibitem{zhu2015aligning}
Y.~Zhu, R.~Kiros, R.~Zemel, R.~Salakhutdinov, R.~Urtasun, A.~Torralba, and
  S.~Fidler, ``Aligning books and movies: Towards story-like visual
  explanations by watching movies and reading books,'' in \emph{Proceedings of
  the IEEE International Conference on Computer Vision}, 2015, pp. 19--27.

\bibitem{fartash2017vse++}
F.~Faghri, D.~J. Fleet, J.~R. Kiros, and S.~Fidler, ``Vse++: Improving
  visual-semantic embeddings with hard negatives,'' \emph{arXiv preprint
  arXiv:1707.05612}, 2017.

\bibitem{yang2018person}
X.~Yang, P.~Zhou, and M.~Wang, ``Person reidentification via structural deep
  metric learning,'' \emph{IEEE transactions on neural networks and learning
  systems}, vol.~30, no.~10, pp. 2987--2998, 2019.

\bibitem{liu2020deep}
X.~Liu, X.~Yang, M.~Wang, and R.~Hong, ``Deep neighborhood component analysis
  for visual similarity modeling,'' \emph{ACM Transactions on Intelligent
  Systems and Technology (TIST)}, vol.~11, no.~3, pp. 1--15, 2020.

\bibitem{dong2021fine}
J.~Dong, Z.~Ma, X.~Mao, X.~Yang, Y.~He, R.~Hong, and S.~Ji, ``Fine-grained
  fashion similarity prediction by attribute-specific embedding learning,''
  \emph{IEEE Transactions on Image Processing}, vol.~30, pp. 8410--8425, 2021.

\bibitem{xie2017aggregated}
S.~Xie, R.~Girshick, P.~Doll{\'a}r, Z.~Tu, and K.~He, ``Aggregated residual
  transformations for deep neural networks,'' in \emph{Proceedings of the
  IEEE/CVF conference on Computer Vision and Pattern Recognition}, 2017, pp.
  1492--1500.

\bibitem{mettes2020shuffled}
P.~Mettes, D.~C. Koelma, and C.~G. Snoek, ``Shuffled imagenet banks for video
  event detection and search,'' \emph{ACM Transactions on Multimedia Computing,
  Communications, and Applications}, vol.~16, no.~2, pp. 1--21, 2020.

\bibitem{cvpr2016-resnet}
K.~He, X.~Zhang, S.~Ren, and J.~Sun, ``Deep residual learning for image
  recognition,'' in \emph{Proceedings of the IEEE/CVF conference on Computer
  Vision and Pattern Recognition}, 2016, pp. 770--778.

\bibitem{li2019w2vv++}
X.~Li, C.~Xu, G.~Yang, Z.~Chen, and J.~Dong, ``W2vv++ fully deep learning for
  ad-hoc video search,'' in \emph{Proceedings of the 27th ACM International
  Conference on Multimedia}, 2019, pp. 1786--1794.

\bibitem{yu2017end}
Y.~Yu, H.~Ko, J.~Choi, and G.~Kim, ``End-to-end concept word detection for
  video captioning, retrieval, and question answering,'' in \emph{Proceedings
  of the IEEE/CVF conference on Computer Vision and Pattern Recognition}, 2017,
  pp. 3165--3173.

\bibitem{luo2020univl}
H.~Luo, L.~Ji, B.~Shi, H.~Huang, N.~Duan, T.~Li, J.~Li, T.~Bharti, and M.~Zhou,
  ``Univl: A unified video and language pre-training model for multimodal
  understanding and generation,'' \emph{arXiv preprint arXiv:2002.06353}, 2020.

\bibitem{patrick2020support}
M.~Patrick, P.-Y. Huang, Y.~Asano, F.~Metze, A.~Hauptmann, J.~Henriques, and
  A.~Vedaldi, ``Support-set bottlenecks for video-text representation
  learning,'' \emph{arXiv preprint arXiv:2010.02824}, 2020.

\bibitem{rouditchenko2020avlnet}
A.~Rouditchenko, A.~Boggust, D.~Harwath, D.~Joshi, S.~Thomas, K.~Audhkhasi,
  R.~Feris, B.~Kingsbury, M.~Picheny, A.~Torralba \emph{et~al.}, ``Avlnet:
  Learning audio-visual language representations from instructional videos,''
  \emph{arXiv preprint arXiv:2006.09199}, 2020.

\bibitem{lei2018tvqa}
J.~Lei, L.~Yu, M.~Bansal, and T.~L. Berg, ``Tvqa: Localized, compositional
  video question answering,'' \emph{arXiv preprint arXiv:1809.01696}, 2018.

\bibitem{pan2020auto}
Y.~Pan, Y.~Li, J.~Luo, J.~Xu, T.~Yao, and T.~Mei, ``Auto-captions on gif: A
  large-scale video-sentence dataset for vision-language pre-training,''
  \emph{arXiv preprint arXiv:2007.02375}, 2020.

\bibitem{iccv2019-francis}
D.~Francis, P.~Anh~Nguyen, B.~Huet, and C.-W. Ngo, ``Fusion of multimodal
  embeddings for ad-hoc video search,'' in \emph{Proceedings of the IEEE/CVF
  International Conference on Computer Vision Workshops}, 2019.

\bibitem{li2020novel}
Z.~Li, C.~Guo, B.~Yang, Z.~Feng, and H.~Zhang, ``A novel convolutional
  architecture for video-text retrieval,'' in \emph{IEEE International
  Conference on Multimedia and Expo}, 2020, pp. 1--6.

\bibitem{zhao2020stacked}
R.~Zhao, K.~Zheng, and Z.-j. Zha, ``Stacked convolutional deep encoding network
  for video-text retrieval,'' in \emph{IEEE International Conference on
  Multimedia and Expo}, 2020, pp. 1--6.

\bibitem{wei2021universal}
J.~Wei, Y.~Yang, X.~Xu, X.~Zhu, and H.~T. Shen, ``Universal weighting metric
  learning for cross-modal retrieval,'' \emph{IEEE Transactions on Pattern
  Analysis and Machine Intelligence}, 2021.

\bibitem{chen2020interclass}
F.~Chen, J.~Shao, Y.~Zhang, X.~Xu, and H.~T. Shen,
  ``Interclass-relativity-adaptive metric learning for cross-modal matching and
  beyond,'' \emph{IEEE Transactions on Multimedia}, 2020.

\bibitem{2015Order}
I.~Vendrov, R.~Kiros, S.~Fidler, and R.~Urtasun, ``Order-embeddings of images
  and language,'' \emph{arXiv preprint arXiv:1511.06361}, 2015.

\bibitem{feng2014cross}
F.~Feng, X.~Wang, and R.~Li, ``Cross-modal retrieval with correspondence
  autoencoder,'' in \emph{Proceedings of the 22nd ACM international conference
  on Multimedia}, 2014, pp. 7--16.

\bibitem{frome2013devise}
A.~Frome, G.~Corrado, J.~Shlens, S.~Bengio, J.~Dean, M.~Ranzato, and
  T.~Mikolov, ``Devise: A deep visual-semantic embedding model,'' in
  \emph{Advances in Neural Information Processing Systems}, 2013.

\bibitem{bain2021frozen}
M.~Bain, A.~Nagrani, G.~Varol, and A.~Zisserman, ``Frozen in time: A joint
  video and image encoder for end-to-end retrieval,'' \emph{arXiv preprint
  arXiv:2104.00650}, 2021.

\bibitem{wray2021semantic}
M.~Wray, H.~Doughty, and D.~Damen, ``On semantic similarity in video
  retrieval,'' in \emph{Proceedings of the IEEE/CVF Conference on Computer
  Vision and Pattern Recognition}, 2021, pp. 3650--3660.

\end{thebibliography}
\end{document}